\documentclass{article}

    \usepackage[preprint]{neurips_2025}

\usepackage[utf8]{inputenc} 
\usepackage[T1]{fontenc}    
\usepackage{hyperref}       
\usepackage{url}            
\usepackage{booktabs}       
\usepackage{amsfonts}       
\usepackage{nicefrac}       
\usepackage{microtype}      
\usepackage{xcolor}         
\usepackage{graphicx}
\usepackage{subfig}
\usepackage{multirow}
\usepackage{multicol}
\usepackage{makecell}
\usepackage{amsmath}
\usepackage{algorithm}
\usepackage{algorithmic}
\usepackage{colortbl}
\usepackage{wrapfig}
\usepackage{cleveref}
\usepackage{enumitem}

\title{Block Rotation is All You Need for MXFP4 Quantization}

\author{%
\textbf{Yuantian Shao}$^{1,2}$ \quad
\textbf{Peisong Wang}$^{2,3}$\thanks{Corresponding author.} \quad
\textbf{Yuanteng Chen}$^{2,3,4}$  \quad
\textbf{Chang Xu}$^{5}$ \\
\textbf{Zhihui Wei}$^{1}$ \quad
\textbf{Jian Cheng}$^{2,3}$ \\
  $^1$Nanjing University of Science and Technology, \\
  $^2$ $\text{C}^2$DL, Institute of Automation, Chinese Academy of Sciences, \\
  $^3$School of Artificial Intelligence, University of Chinese Academy of Sciences, \\
  $^4$Zhongguancun Academy, \\
  $^5$School of Computer Science, University of Sydney. \\
}

\begin{document}

\maketitle

\begin{abstract}
Large language models (LLMs) have achieved remarkable success, but their rapidly growing scale imposes prohibitive costs in memory, computation, and energy. Post-training quantization (PTQ) is a promising solution for efficient deployment, yet achieving accurate W4A4 quantization remains an open challenge. While most existing methods are designed for INT4 formats, the emergence of MXFP4—a new FP4 format with various hardware support (NVIDIA, AMD, Intel)—raises questions about the applicability of current techniques. In this work, we establish a comprehensive benchmark of PTQ methods under the MXFP4 format. Through systematic evaluation, we find that methods like GPTQ consistently deliver strong performance, whereas rotation-based approaches, which are almost used by all state-of-the-art approaches, suffer from severe incompatibility with MXFP4. We further provide the first in-depth analysis of this conflict, tracing its root to a fundamental mismatch between MXFP4’s PoT (power-of-two) block scaling and the redistribution of outlier energy via global rotation. Building on this insight, we propose a simple yet effective block rotation strategy that adapts rotation-based methods to MXFP4, leading to substantial accuracy improvements across diverse LLMs. Our findings not only offer clear guidance for practitioners but also set a foundation for advancing PTQ research under emerging low-precision formats.
\end{abstract}

\section{Introduction}
Large language models (LLMs) have become the cornerstone of modern artificial intelligence, but their ever-increasing scale incurs substantial memory, computation, and energy costs~\citep{wu2025survey,dantas2025review}. Among numerous model compression techniques, post-training quantization (PTQ) has emerged as a practical solution due to its training-free nature and low engineering overhead~\citep{czako2025addressing}. While INT8 and INT4 quantization have already been adopted in practice, achieving accurate W4A4 (4-bit weights and 4-bit activations) remains a critical challenge~\citep{elangovan2025bcq}. For more recently deliberated LLM models (e.g., LLaMA-3.2 1B/3B), naive 4-bit quantization often results in severe performance degradation~\citep{van2025fptquant}, making W4A4 a key research frontier for efficient LLM deployment.

Meanwhile, hardware advances have spurred the microscaling (MX) family of data formats~\citep{han2025bbal}, such as MXFP4. MXFP4 is an open standard format proposed by the Open Compute Project (OCP)~\cite{ocp_mx_spec_2023}, and currently has been supported by AMD Ryzen AI MAX+ 395~\citep{amd2025gptoss}, Nvidia RTX 5090/B200~\citep{tensorrt_llm}, etc. Compared with INT4, FP4 is better suited to handling long-tailed distributions~\citep{lee2024amxfp4}. The use of shared block-scale factors extends the representable dynamic range while simultaneously restricting the influence of outliers. It supports not only inference but also low-precision training~\citep{amd_quark,microxcaling}, and can be efficiently emulated or converted on diverse platforms, including Apple M-series chips, NVIDIA Ampere/Ada GPUs, and common x86 CPUs, thus offering broader software and hardware compatibility. To our knowledge, the model openai/gpt-oss~\citep{openai_gpt_oss}, as the first LLM with native FP4 support, adopts MXFP4, underscoring its importance among future low-precision formats.

Existing W4A4 methods are primarily designed for INT4 quantization and are typically evaluated under different datasets, quantization settings, or simulation modes. As a result, practitioners lack clear guidance on how to apply these methods to the MXFP4 format. To address this gap, we categorize existing PTQ methods into three groups: (1) compensation-based, (2) transformation-based, and (3) optimization-based. We then conduct a detailed comparative analysis of representative methods within each category under the MXFP4 format. Our evaluation highlights methods that achieve significant improvements, and reveals the incompatibility between rotation-based techniques and MXFP4.

Furthermore, we investigate why combining rotation with MXFP4 leads to performance collapse~\citep{lee2024amxfp4}. To the best of our knowledge, this is the first in-depth study of this issue. We attribute the root cause to a fundamental mismatch: MXFP4 uses a shared PoT (power-of-two) block-scale mechanism to suppress outliers, whereas rotation methods attempt to mitigate them by distributing their energy across all channels. Based on this insight, we propose a grouped rotation strategy to adapt rotation-based methods to MXFP4. This strategy can be easily integrated into existing rotation schemes and substantially improves PTQ accuracy under MXFP4. Our work not only provides practitioners with clear guidance for selecting effective quantization methods but also establishes a direction for further community efforts in optimizing MXFP4 PTQ.

The key contributions of this paper are summarized as follows:

\begin{itemize}[leftmargin=12pt,topsep=0pt,partopsep=0pt]
\item We established a W4A4 quantization benchmark for the MXFP4 format, systematically categorized existing PTQ methods, conducted a detailed evaluation of representative approaches, and highlighted their limitations under this new format.

\item We conduct a thorough investigation of rotation-based methods under the MXFP4 format, identifying that the destructive interaction is fundamentally caused by the combination of PoT scales failing to recover large values within blocks and global rotations amplifying originally small values.

\item Building on this insight, we propose a Block-wise Rotation Quantization (BRQ) strategy that adapts rotation methods to MXFP4. This strategy can be seamlessly integrated into existing rotation schemes and substantially improves PTQ accuracy under MXFP4 across multiple models and tasks.
\end{itemize}

\section{Categorization of PTQ Methods}\label{sec:ctg}

We focus on fully quantized W4A4 PTQ methods, excluding mixed-precision schemes to ensure fair and consistent evaluation under MXFP4. Our benchmark systematically examines existing low-bit PTQ approaches for LLMs in this setting. For clarity, we categorize them into three classes: compensation-based, transformation-based, and optimization-based.

\subsection{Compensation-Based Quantization Methods}

Compensation-based methods reduce quantization errors by adjusting quantized weights to correct low-bit perturbations. GPTQ~\citep{frantar2022gptq}, a representative approach, which performs column-wise offline optimization of weight matrices utilizing second-order information approximations from the Hessian matrix, achieving precise compensation and significantly reducing overall quantization loss. Subsequent methods extend this principle: BoA~\citep{kim2024boa} incorporates attention-aware Hessians, RSQ~\citep{sung2025rsq} applies token-wise weighting, QuantEase~\citep{behdin2023quantease} leverages coordinate descent for forward reconstruction, VPTQ~\citep{liu2024vptq} combines vector quantization with channel-independent second-order optimization, and APTQ~\citep{guan2024aptq} uses Hessian traces to guide selective mixed-precision quantization.

Together, these compensation-based approaches share the principle of explicit error correction, making them particularly effective for transformer-based LLMs, especially in attention-dense modules sensitive to low-bit perturbations.

\subsection{Transformation-Based Methods}

In the low-bit case, outliers can significantly increase the quantization error. Applying carefully designed equivalent transformations can redistribute or reshape the data to reduce the impact of extreme values. SmoothQuant~\citep{xiao2023smoothquant} applies a smoothing transformation to redistribute large activation outliers to the corresponding weight scales, thereby mitigating their impact on low-bit quantization. Building on a similar principle, QServe~\citep{lin2024qserve} integrates progressive low-bit quantization with system-level optimization and SmoothAttention to improve inference throughput while maintaining model fidelity. QuIP~\cite{chee2024quip} introduces incoherent processing to decorrelate the contributions of outliers in both weight and activation spaces. QuIP$\#$~\cite{tseng2024quip} further enhances computational efficiency by employing a randomized Hadamard transform, which improves orthogonality and reduces inter-channel coherence. QuaRot~\citep{ashkboos2024quarot} and DuQuant~\citep{lin2024duquant} leverage rotation transforms to spread outlier values across subspaces of smaller-magnitude activations or multiple channels, reducing sensitivity to low-bit representation and improving reconstruction accuracy.

These transformation-based methods are particularly effective for modules that exhibit high activation variance or extreme outliers and are fundamental to optimization-based methods.

\subsection{Optimization-Based Methods}

Given the difficulty of manually designing equivalent transformations, some studies propose parameterizing these transformations as learnable variables, allowing them to be optimized within the model to achieve higher performance. OmniQuant~\citep{shao2023omniquant} introduces learnable weight clipping and equivalent transformations to achieve superior W4A4 quantization performance. SpinQuant~\citep{liu2024spinquant} demonstrates that optimizing rotation matrices is more effective than random transformations in dispersing weight outliers, significantly reducing quantization errors in extremely low-bit scenarios. AffineQuant~\citep{ma2024affinequant} and FlatQuant~\citep{sun2024flatquant} extend this principle by applying affine transformations to jointly adjust weights and activations, flattening distributions to mitigate the impact of outliers and simplify the optimization process. KurTail~\citep{sadegh2025kurtail} leverages kurtosis-based rotation to alleviate outliers in LLM activations, achieving high-fidelity low-bit quantization.

Overall, optimization-driven methods can fully exploit gradient information to adaptively adjust weights and activations under strict low-bit constraints, achieving near-optimal accuracy in low-bit settings.

\section{Benchmark and Analyze} \label{sec:benchmark}

To assess whether existing PTQ algorithms fully exploit MXFP4, we establish a comprehensive benchmark. Instead of proposing new techniques, our aim is to objectively evaluate representative methods from three categories in this section. We test seven state-of-the-art approaches on models of varying scales. To capture overall trends, we average perplexity and downstream accuracy across models and report results in Figure~\ref{fig:benchmark}. This benchmark provides a fair basis for comparison and reveals key limitations that motivate the existing method.

\subsection{Experimental Setup}
All experiments are carried out on NVIDIA A800 GPU servers, with MX format quantization simulated using Microsoft’s open-source repository \textit{microsoft/microxcaling}~\citep{microxcaling}.

\textbf{Method selection.} We selected representative PTQ methods from the three categories in Section~\ref{sec:ctg}. For compensation-based methods, we chose GPTQ~\citep{frantar2022gptq}. Transformation-based methods include SmoothQuant~\citep{xiao2023smoothquant} and QuaRot~\citep{ashkboos2024quarot}. Notably, we distinguish between two variants of QuaRot: QuaRot, which applies random rotation with RTN, and QuaRot+, which integrates random rotation with GPTQ, in order to separately evaluate the effect of random rotation alone and in combination with GPTQ. Optimization-based methods comprise OmniQuant~\citep{shao2023omniquant} and SpinQuant~\citep{liu2024spinquant}, representing parameter optimization and end-to-end rotation optimization, respectively. We also include round-to-nearest (RTN) INT4 with block size 32 and FP16 scale as a naive baseline (BINT4).

\textbf{Models.} We benchmarked selected methods on multiple widely adopted large language models, including LLaMA-2 7B/13B, LLaMA-3 8B, LLaMA-3.2 1B/3B and Mistral-7B, which span different scales and architectures. These models represent a spectrum of modern transformer-based LLMs and provide a robust testbed for quantization research.

\textbf{Datasets and metrics}. We evaluated perplexity (PPL) on WikiText2 as a proxy for language modeling quality, and accuracy on five zero-shot downstream tasks: PIQA \citep{bisk2020piqa}, WinoGrande \citep{sakaguchi2021winogrande}, OpenBookQA \cite{mihaylov2018can}, ARC-Easy and ARC-Challenge \citep{boratko2018systematic}.

\subsection{Evaluations}
As shown in Table~\ref{tab:benchmark}, MXFP4 RTN suffers a substantial accuracy drop compared to FP16 and even BINT4 baselines. This demonstrates that despite MXFP4's significant hardware advantages, PTQ on it remains a significant challenge, further highlighting the need to systematically evaluate the performance of existing PTQ methods on MXFP4.

\begin{table*}[!t]
\setlength{\tabcolsep}{2pt} 
\small
\centering
\caption{Comparison of WikiText perplexity (Wiki) and average zero-shot accuracy (Avg.) across multiple LLMs under FP16, BINT4, and MXFP4 quantization. QuaRot$^+$ denotes the variant integrated with the GPTQ algorithm. \textbf{The best results are highlighted in black bold, while the worst results are highlighted in gray bold.} Detailed results are provided in Appendix~\ref{apx:results}.}
\label{tab:benchmark}
\begin{tabular}{c|cc|cc|cc|cc|cc|cc}
\hline
\hline
\multirow{2}{*}{Method} & \multicolumn{2}{c|}{LLaMA-2 7B} & \multicolumn{2}{c|}{LLaMA-2 13B} & \multicolumn{2}{c|}{LLaMA-3 8B} & \multicolumn{2}{c|}{LLaMA-3.2 1B} & \multicolumn{2}{c|}{LLaMA-3.2 3B} & \multicolumn{2}{c}{Mistral 7B} \\
& Wiki & Avg. & Wiki & Avg. & Wiki & Avg. & Wiki & Avg. & Wiki & Avg. & Wiki & Avg. \\
\hline
FP16 & 5.47 & 62.59 & 4.88 & 64.89 & 6.14 & 65.85 & 9.75 & 53.81 & 7.81 & 61.60 & 5.25 & 66.73 \\
BINT4 & 5.94 & 61.30 & 5.16 & 63.32 & 7.40 & 63.12 & 13.56 & 48.36 & 9.29 & 57.47 & 5.63 & 65.28 \\
\hline
RTN & 7.08 & 57.26 & 5.90 & 61.40 & 8.23 & 60.61 & 15.91 & 46.89 & 10.27 & 55.22 & 6.56 & 62.86 \\
GPTQ & 6.56 & \textbf{59.27} & 5.41 & 62.91 & 7.68 & 61.48 & 13.35 & 48.52 & \textbf{9.50} & 55.40 & 6.00 & 63.34 \\
SmoothQuant & 7.04 & 57.18 & 5.73 & 61.52 & 8.11 & 61.22 & 16.86 & 46.48 & 10.38 & 55.05 & 6.49 & 62.96 \\
QuaRot & \textbf{\textcolor{gray}{13.09}} & \textbf{\textcolor{gray}{50.32}} & \textbf{\textcolor{gray}{7.03}} & \textbf{\textcolor{gray}{59.09}} & \textbf{\textcolor{gray}{9.56}} & \textbf{\textcolor{gray}{59.26}} & \textbf{\textcolor{gray}{17.86}} & \textbf{\textcolor{gray}{45.42}} & \textbf{\textcolor{gray}{13.36}} & \textbf{\textcolor{gray}{51.60}} & \textbf{\textcolor{gray}{6.65}} & \textbf{\textcolor{gray}{60.33}} \\
QuaRot$^+$ & 6.29 & 58.35 & 5.57 & 61.57 & 7.68 & 61.57 & 12.78 & 48.83 & 9.92 & 55.91 & 5.73 & 63.66 \\
OmniQuant & 6.56 & 56.67 & 5.43 & 61.89 & 8.16 & 60.47 & 14.32 & 48.17 & 9.85 & 55.76 & 6.37 & 61.82 \\
SpinQuant & \textbf{5.99} & 59.24 & \textbf{5.20} & \textbf{62.78} & \textbf{7.62} & \textbf{61.93} & \textbf{12.72} & \textbf{49.09} & 9.85 & \textbf{56.19} & \textbf{5.68} & \textbf{63.79} \\
\hline
\hline
\end{tabular}
\vskip -2em
\end{table*}

In addition, we can see that most methods yield some improvements when directly applied to the MXFP4 format. GPTQ stands out by consistently delivering notable gains, even surpassing BINT4 on certain models (e.g., LLaMA-3.2 1B: 15.91/46.89 → 13.35/48.52). However, other methods are less reliable. OmniQuant requires delicate hyperparameter tuning to achieve stable optimization on small models (LLaMA-3.2 1B/3B), yet still underperforms GPTQ (e.g., LLaMA 3.2 1B PPL 14.32 vs. 3.35). SmoothQuant provides only marginal benefits and can even harm performance, revealing MXFP4’s heightened sensitivity to parameter magnitudes under its scaling scheme. 

The most striking results arise from rotation-based methods. QuaRot, when combined with RTN, leads to catastrophic degradation (e.g., LLaMA-2 7B: 7.08/57.26 → 13.09/50.32). Even when integrated with GPTQ, performance gains remain inconsistent and limited. This indicates a structural incompatibility between random rotations and MXFP4, but the root cause of this destructive interaction has not yet been thoroughly discussed by research~\citep{lee2024amxfp4}. SpinQuant leverages a straight-through estimator for end-to-end optimization, which enables rotations to adaptively align with the non-uniform scaling of MXFP4. While this enforced optimization does alleviate the incompatibility to some extent, it delivers only marginal improvements over QuaRot$^+$ (e.g., Mistral 7B: 5.73/63.66 → 5.68/63.79), suggesting that optimization alone does not fully resolve the compatibility issue.

Given the critical role of rotation in INT4 quantization, we further examine its impact across other commonly used quantization formats. Figure~\ref{fig:format} reports results for rotation and its variants under INT4 (weight per-channel symmetric quantization with per-token asymmetric activation quantization), BINT4, BFP4 (FP4 variant of BINT4), as well as MXINT4 and MXFP4. The key findings can be summarized as follows:

\begin{figure}[!t]
    \centering
    \begin{minipage}{0.49\linewidth}
        \centering
        \includegraphics[width=0.9\linewidth]{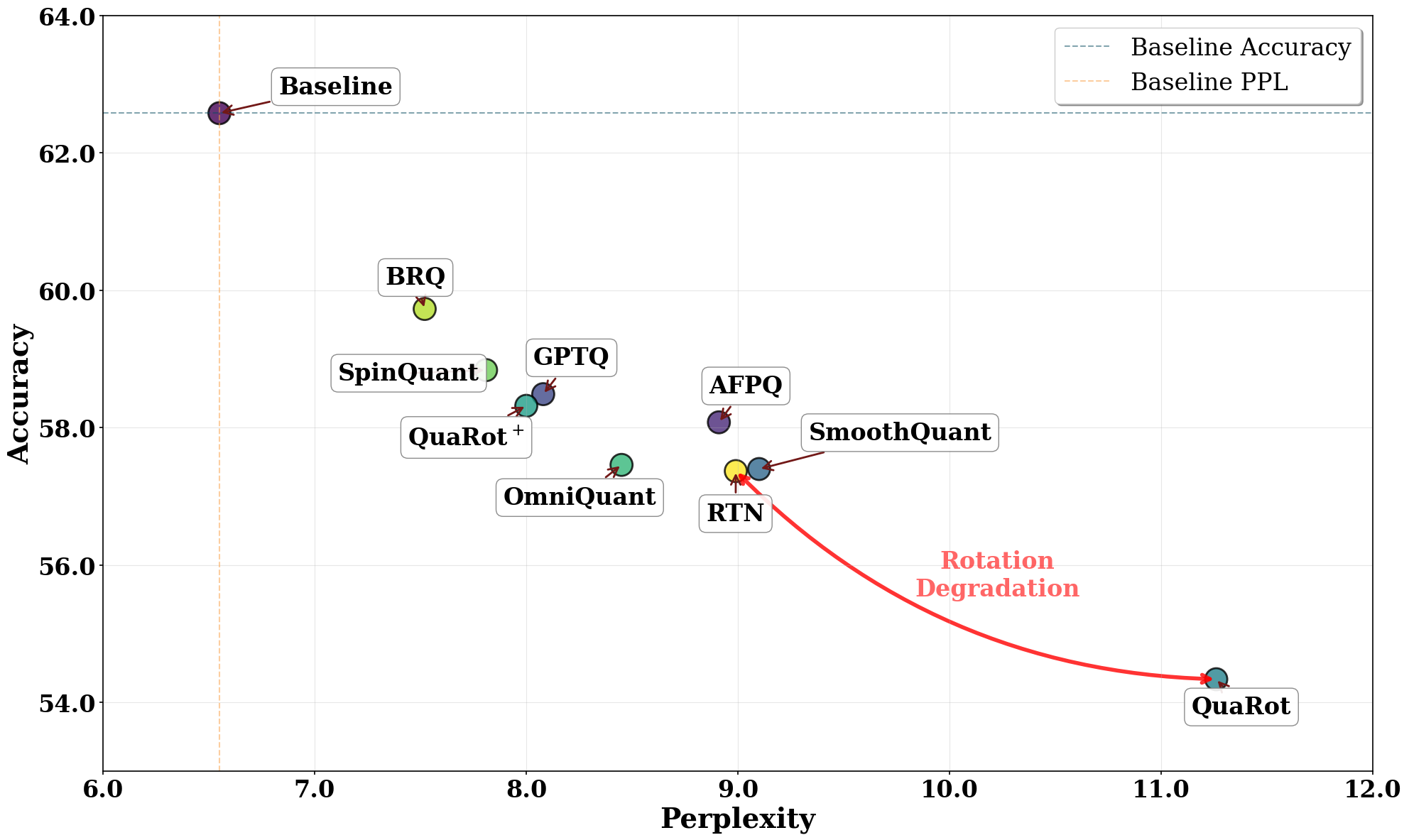}
        \caption{Overall performance of quantization methods under MXFP4. The x-axis shows perplexity, the y-axis shows average downstream accuracy, and methods nearer the top-left are closer to the FP16 baseline, indicating better performance.}
        \label{fig:benchmark}
    \end{minipage}
    \hfill
    \begin{minipage}{0.49\linewidth}
        \centering
        \includegraphics[width=0.9\linewidth]{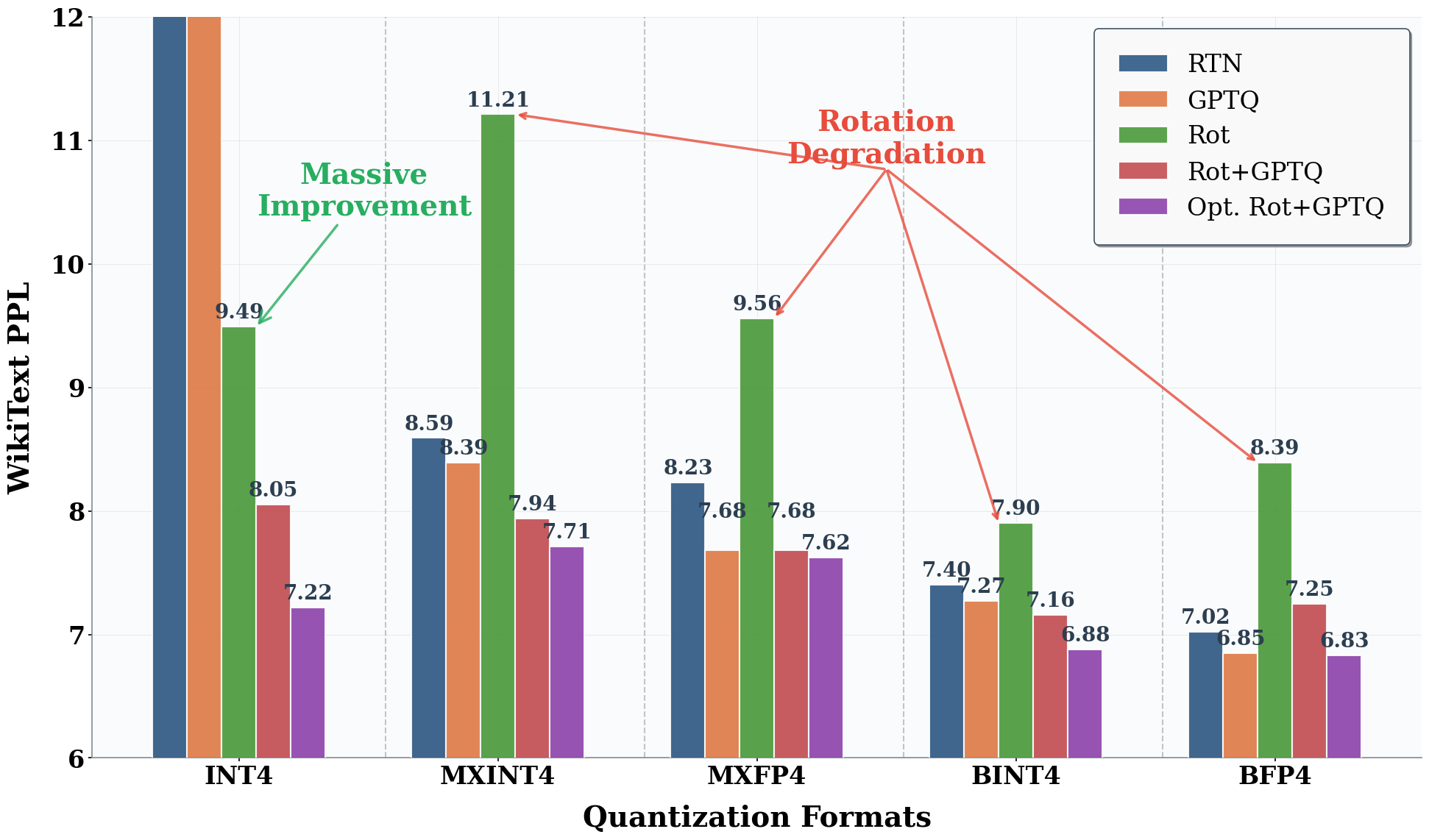}
        \caption{Effect of rotation and its variants across different quantization formats. Rot applies a random Hadamard transform with RTN; Rot+GPTQ combines the transform with GPTQ; and Opt. Rot+GPTQ employs an optimized rotation matrix with GPTQ.}
        \label{fig:format}
    \end{minipage}
    \vskip -2em
\end{figure}

\textbf{A. INT4 benefits substantially from rotation.} In the widely studied INT4 setting, applying rotation alone yields significant performance improvements. When combined with GPTQ or rotation optimization, the gains are further amplified, indicating that rotation is particularly effective for uniformly distributed integer formats.

\textbf{B. FP4 formats outperform INT4 without rotation.} When not using rotation, BFP4 and MXFP4 achieve consistently higher performance than BINT4 and MXINT4, suggesting that FP4’s wider dynamic range and representational flexibility are better suited for 4-bit quantization.

\textbf{C. Random rotation degrades performance in group-wise formats.} In contrast to INT4, group-wise quantization formats (BINT4, BFP4, MXINT4, MXFP4) suffer from performance degradation under random rotation. The effect is especially pronounced in MX-based formats, where performance can drop below that of simple RTN.

\textbf{D. Divergent behaviors under FP16 vs. PoT scaling.} For FP16-scale formats, BINT4 outperforms BFP4 after random rotation. Conversely, for PoT-scale formats, MXFP4 underperforms compared to its FP16-scale counterpart, with rotation amplifying the discrepancy.

\textbf{E. PoT scaling in MX formats incurs additional loss.} Comparing MXINT4/MXFP4 against their FP16-scale counterparts BINT4/BFP4, PoT scaling consistently introduces larger quantization errors, which become even more severe after rotation.

\textbf{F. Optimized rotation remains limited on MXFP4.} While optimized rotation combined with GPTQ improves MXFP4 performance, the final results still lag behind those of INT4 under comparable configurations.

Overall, these results reveal a striking divergence: while rotation and its variants consistently enhances INT4, it fails to generalize to MXFP4. In particular, the interaction between rotation and MXFP4’s block-wise PoT scaling leads to unique degradation patterns. This raises an important open question: \textbf{Why does a technique that is fundamentally beneficial in INT4 become harmful in MXFP4?} To address this, we next conduct a deeper analysis of MXFP4’s structural characteristics and their destructive interplay with rotation-based transformations.

\section{Why Rotation Transforms Hurt MXFP4} \label{sec:block_rotation}

To understand why rotation transformations—despite their remarkable success on INT4—degrade quantization accuracy under the MXFP4 format, we first dissect the unique characteristics and inherent limitations of MXFP4. We then analyze how rotation reshapes model data distributions, and by synthesizing these perspectives, we identify the root cause of their conflict. Building on this insight, we further propose a practical solution to reconcile the incompatibility.

\subsection{Limited Recovery of Large Values in MXFP4 Blocks}

MXFP4 represents each value in the E2M1 format, with one sign bit, one mantissa bit, and two exponent bits. It applies symmetric per-group quantization with a fixed group size of 32, where each group is associated with an E8M0 (PoT) scaling factor directly integrated into hardware. With finer granularity and FP4’s non-uniform representation, MXFP4 achieves a substantially higher quantization signal-to-noise ratio (QSNR) than per-tensor or per-channel INT4, thus better approximating full-precision values~\citep{darvish2023shared}.

\begin{figure*}[!t]
\begin{center}
\subfloat[PoT round error.]{\includegraphics[width=0.32\linewidth]{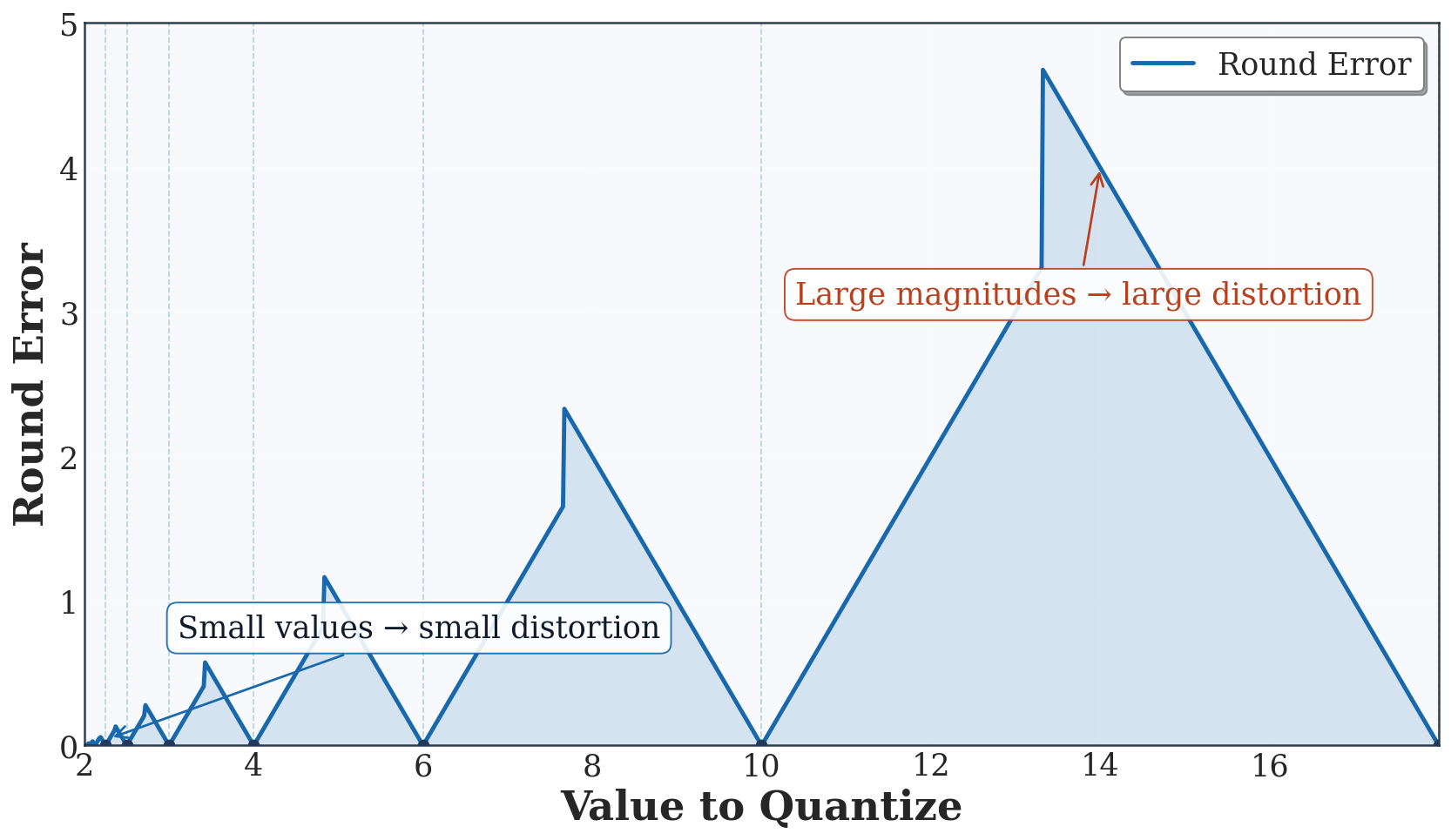}\label{fig: pot}} 
\subfloat[Regular block.]{\includegraphics[width=0.32\linewidth]{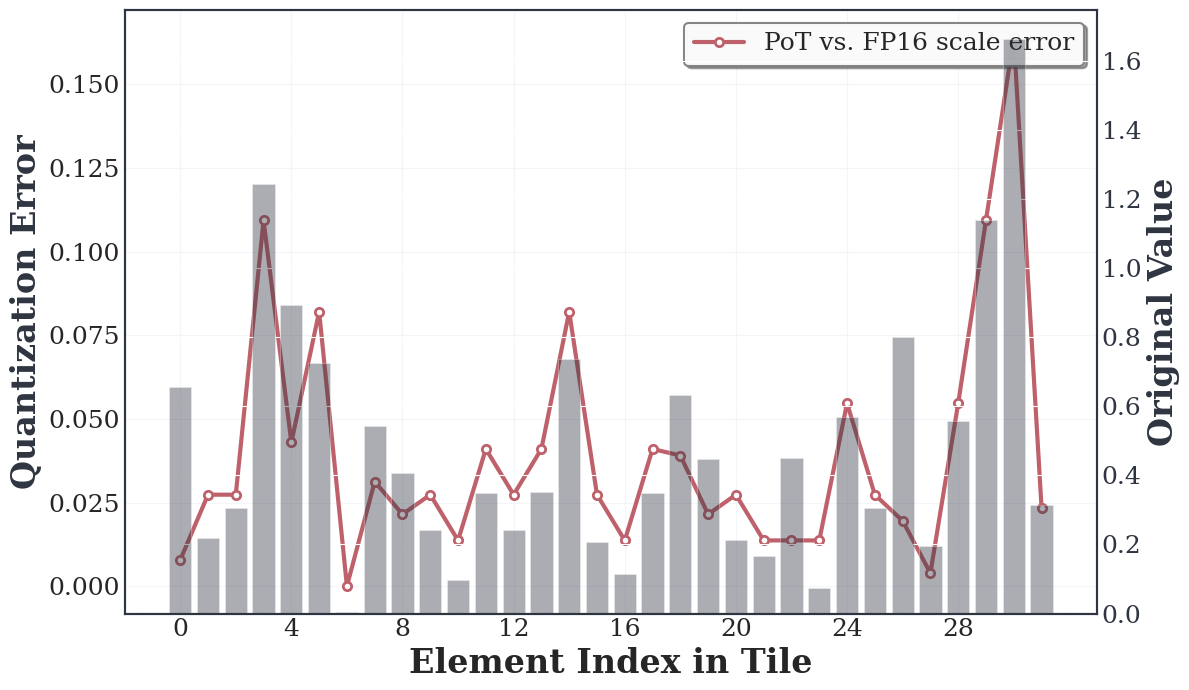}\label{fig: regular}}
\subfloat[Outlier block.]{\includegraphics[width=0.32\linewidth]{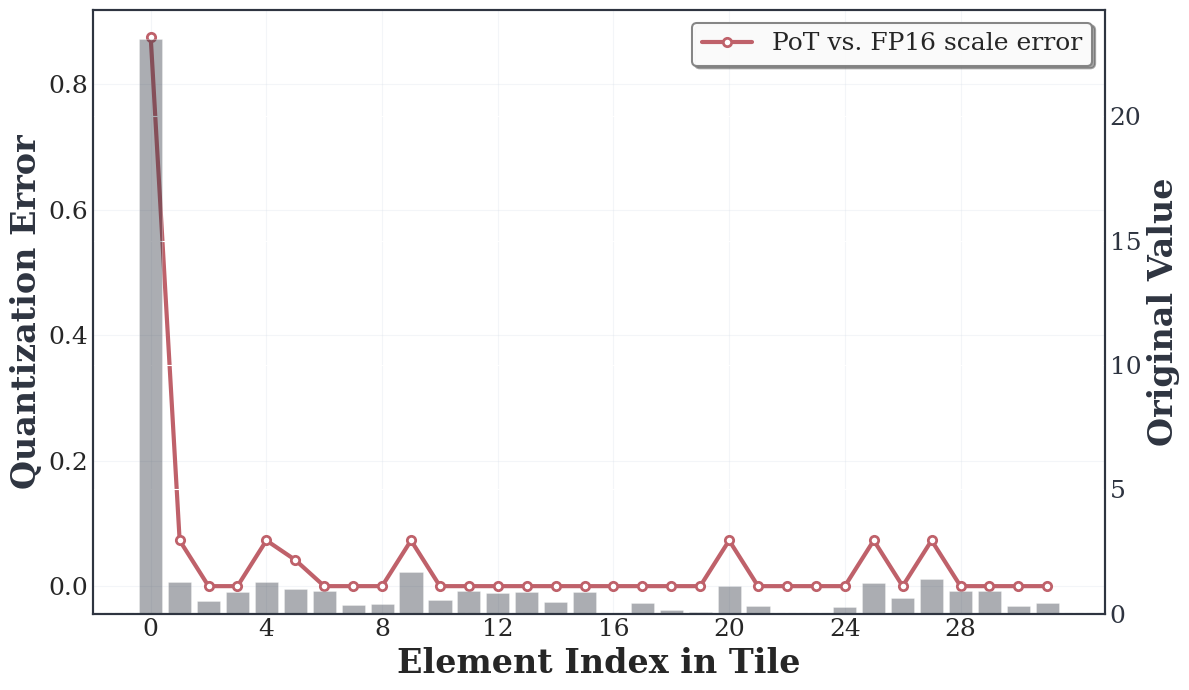}\label{fig: outlier}}
\caption{(a) illustrates the rounding error curve of PoT format. (b) and (c) show the quantization error of MXFP4 relative to BFP4 for regular and outlier blocks, respectively. Bar charts represent the original activation values (right axis), lines indicate the relative quantization error (left axis).}
\label{fig:diff_scale}
\end{center}
\vskip -2em
\end{figure*}

Figure~\ref{fig:format} shows that BFP4 and MXFP4 can lead to significant performance differences due to different scale formats. To further investigate the differences between these scales, we categorize quantization blocks into two types: “\textbf{regular blocks},” which contain no outliers, and “\textbf{outlier blocks},” which include one or more outliers. Figure~\ref{fig: regular} and~\ref{fig: outlier} visualizes MXFP4’s quantization error relative to BFP4 for both block types. We observe that for both regular and outlier blocks, the error increases with the magnitude of the elements. Notably, in outlier blocks, the quantization loss of outlier is up to five times larger than the maximum loss in regular blocks. This is primarily due to the PoT format’s coarse granularity at large magnitudes, which amplifies rounding errors when representing large values (see Figure~\ref{fig: pot}).

In summary, the main bottleneck of MXFP4 lies in its limited ability to reconstruct large values in blocks, with the reconstruction error increasing sharply with magnitude. Therefore, improving MXFP4 performance ultimately depends on effectively reducing these large values.

\subsection{Rotation Induced Growth of Small Values}
In W4A4 quantization, the primary source of performance degradation is the quantization error of activations~\citep{ashkboos2024quarot}. To investigate the compatibility issues between rotation-based transformations and the MXFP4 format, we conducted a detailed analysis of activation distributions before and after rotation. 

\begin{figure*}[ht]
\begin{center}
\centerline{\includegraphics[width=\textwidth]{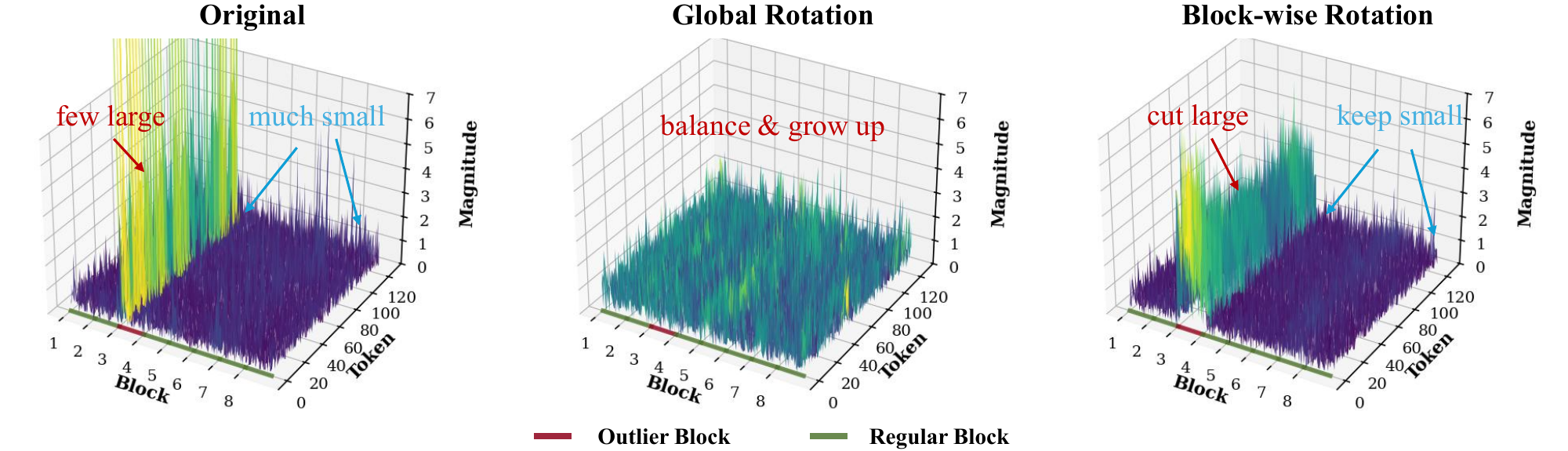}}
\caption{Comparison of the distribution of Llama-3 8B activation after different transformations. More block-scale visualizations are provided in \textbf{Appendix~\ref{apx:rot}}.}
\label{fig:block_rotation}
\end{center}
\vskip -2em
\end{figure*}

\begin{figure}[!t]
    \centering
    \begin{minipage}{0.49\linewidth}
        \centering
        \includegraphics[width=0.85\linewidth]{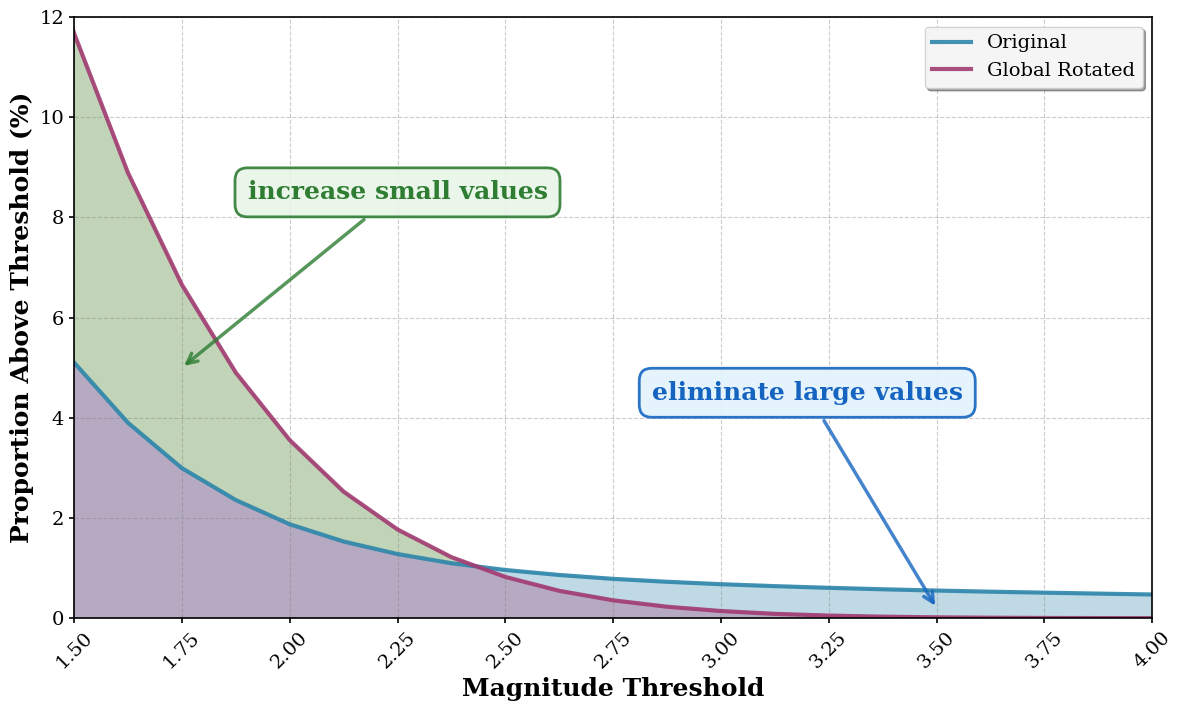}
        \caption{The effect of rotation transformation on activation distribution. The horizontal axis represents the segmentation threshold, and the vertical axis represents the percentage of data greater than the threshold.
        }
        \label{fig:chan by rot}
    \end{minipage}
    \hfill
    \begin{minipage}{0.49\linewidth}
        \centering
        \includegraphics[width=0.85\linewidth]{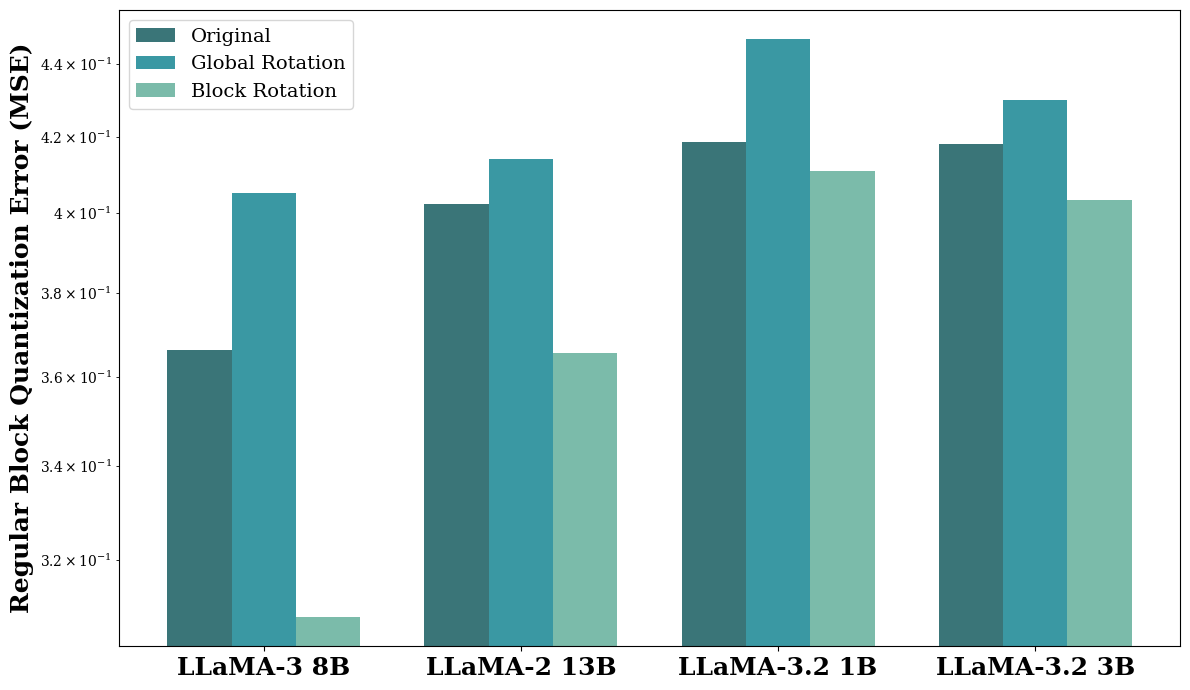}
        \caption{Average quantization loss (logarithmic result) of regular blocks after applying different rotations, where outliers are defined as the top 0.1\% of activations in descending order of absolute value~\citep{dettmers2022gpt3}.
        }
        \label{fig:regular_block_loss}
    \end{minipage}
    \vskip -2em
\end{figure}

As illustrated in Figure~\ref{fig:block_rotation}, conventional rotation methods employ rotation matrices to redistribute outliers originally concentrated in a few channels across all dimensions, thereby reducing quantization error. However, rotation does not reduce the overall energy; the L2 norm of the activations remains unchanged. In effect, the energy from the original outlier channels is not eliminated but redistributed across previously small-value channels, which consequently become magnified. To examine this effect, we sampled 2,048 activations from LLaMA-3 8B and analyzed their distributional shifts after rotation, as shown in Figure~\ref{fig:chan by rot}. The results indicate that rotation largely removes the $\sim$1\% of activations exceeding 3 (corresponding to the blue area in the figure), but at the cost of substantially \textbf{increasing the proportion of activations greater than 1.5} (from about 5\% before rotation to 11\% after rotation, corresponding to the green area in the figure). This evidence clearly demonstrates the effect of rotation on \textbf{amplifying small-value blocks}.

Synthesizing the above observations, we attribute the incompatibility between rotation and MXFP4 quantization to the following destructive interactions:

\begin{itemize}[leftmargin=12pt,topsep=0pt,partopsep=0pt]
\item Global rotation amplifies the scales of regular blocks, thereby increasing their quantization difficulty.

\item The poor reconstruction of large values within MXFP4 blocks further exacerbates the quantization error of these amplified regular blocks.

\item Since regular blocks vastly outnumber outlier blocks, the accumulated errors across them dominate, ultimately leading to a substantial increase in overall quantization loss after rotation.
\end{itemize}

To validate this inference, we measure the average quantization error of regular blocks across different models before and after applying global rotation. As shown in Figure~\ref{fig:regular_block_loss}, regular-blocks' quantization losses significantly increase after rotation, providing strong evidence for our conjecture. Since regular blocks vastly outnumber outlier blocks, this imbalance ultimately leads to the collapse of quantization accuracy under MXFP4 when global rotation is applied.

\subsection{Fix Rotation in MXFP4}

\begin{wrapfigure}{r}{0.5\linewidth}
    \vskip -3em
    \centering
    \includegraphics[width=0.99\linewidth]{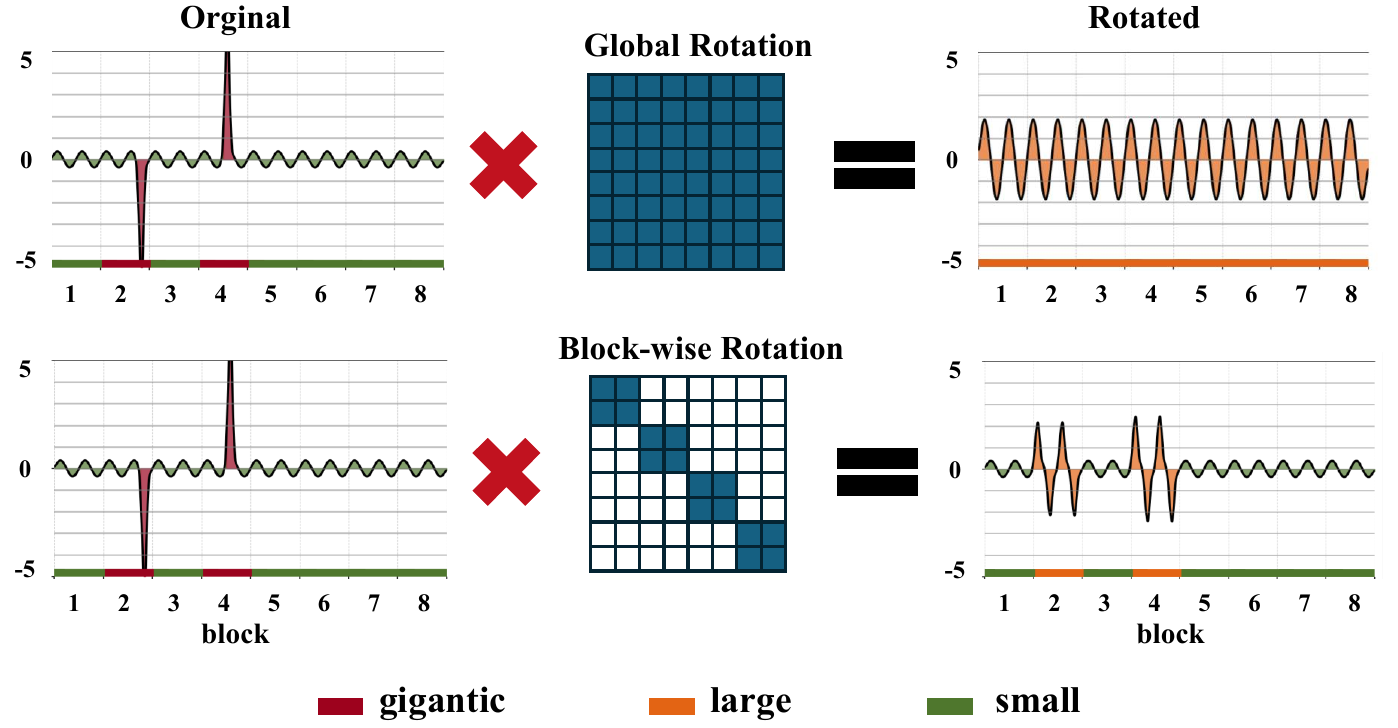}
    \caption{Block rotation’s intuition: Global rotations spread outliers across all channels, inflating regular block scales and worsening quantization error. Block-wise rotations redistribute outliers locally, mitigating outlier effect while keeping regular block scales intact, thereby minimizing quantization error.}
    \label{fig: shiyi}
    \vskip -1em
\end{wrapfigure}

To mitigate the incompatibility between rotation and the block-wise PoT scale inherent to MXFP4, we propose a block-wise rotation strategy, which applies rotation transformations independently within each quantization block, as shown in Figure~\ref{fig: shiyi}. Unlike global rotation, which mixes outliers across all channels, block-wise rotation partitions activations into fixed-size groups (aligned with MXFP4 blocks, e.g., 32 channels) and performs an independent orthogonal transformation within each group. This design preserves the denoising effect of rotation while preventing excessive amplification of small-value channels caused by global mixing.

Formally, let $x \in \mathbb{R}^N$ denote the activation vector for the linear, partitioned into $B$ blocks of size $g$ ($N = B \times g$). Block-wise rotation constructs a block-diagonal matrix:

\begin{equation}
    R_{\mathrm{b l o c k}}=\mathrm{d i a g} ( R_{1}, R_{2}, \ldots, R_{B} ), \quad R_{i} \in\mathbb{R}^{g \times g}, \; R_{i}^{\top} R_{i}=I.
\end{equation}

Block-wise rotation offers multiple advantages over global rotation:

\begin{itemize}[leftmargin=12pt,topsep=0pt,partopsep=0pt]
    \item \textbf{Outlier suppression:} By applying rotations independently within each block, the block rotation matrix effectively redistributes outliers that were previously concentrated in a few channels. This discretization prevents any single outlier from dominating the quantization process, thereby reducing the quantization error of outlier blocks. 

    \item \textbf{Controlled quantization error:} Because each block is rotated independently, the amplification of small-value channels caused by rotation is confined within the block itself. This prevents error propagation across blocks, remains controlled the quantization loss in regular blocks,  which is a key issue in global rotation. 

    \item \textbf{Reduced online computing:} Block-wise rotation substantially reduces the online computation of $R_4$ before the $down_{proj}$ layer. For an input dimension of $N$, global rotation incurs $O(N^2)$ complexity, whereas block-wise rotation reduces it to $O(N \times 32)$. This reduction not only enhances computational efficiency but also facilitates the practical deployment of rotation-based quantization at scale.
\end{itemize}

\section{Experiment}
In this section, we evaluate BRQ using the same test sets and hyperparameters as in Section~\ref{sec:benchmark}. Comparisons are made with GPTQ, QuaRot$^+$, SpinQuant, RTN, and BINT4. Beyond the LLaMA and Mistral models, we also include evaluations on the Qwen~\citep{team2024qwen2} model.

\subsection{Main Results}

\begin{table*}[ht]
\centering
\caption{Performance comparison of BRQ using randomized block rotations and existing PTQ methods without optimization. LLaMA-2 7B/13B/70B and Qwen2.5 7B results can be found in the Appendix~\ref{apx:vs70b}/~\ref{apx:results}.}
\resizebox{\textwidth}{!}{
\begin{tabular}{c|cc|cc|cc|cc|cc|cc}
\hline
\hline
\multirow{2}{*}{Method} & \multicolumn{2}{c|}{LLaMA 3 8B} & \multicolumn{2}{c|}{LLaMA 3.2 1B} & \multicolumn{2}{c|}{LLaMA 3.2 3B} & \multicolumn{2}{c|}{Mistral 7B} & \multicolumn{2}{c|}{Qwen2.5 1.5B} & \multicolumn{2}{c}{Qwen2.5 3B} \\
 & Wiki & Avg & Wiki & Avg & Wiki & Avg & Wiki & Avg & Wiki & Avg & Wiki & Avg \\
\hline
FP16 & 6.14 & 65.85 & 9.75 & 53.81 & 7.81 & 61.60 & 5.25 & 66.73 & 9.87 & 58.83 & 8.03 & 61.91 \\
BINT4 & 7.40 & 63.12 & 13.56 & 48.36 & 9.29 & 57.47 & 5.63 & 65.28 & 13.98 & 53.98 & 10.32	& 58.03 \\
\hline
RTN & 8.23 & 60.61 & 15.91 & 46.89 & 10.27 & 55.22 & 6.56 & 62.86 & 16.61 & 52.69 & 11.03 & 57.77 \\
GPTQ & 7.68 & 61.48 & 13.35 & 48.52 & 9.50 & 55.40 & 6.00 & 63.34 & 13.94 & 53.13 & 10.20 & 58.23 \\
QuaRot$^+$ & 7.68 & 61.57 & 12.78 & 48.83 & 9.92 & 55.91 & 5.73 & 63.66 & 12.80 & 53.50 & 9.65 & 58.27 \\
BRQ & \textbf{7.14} & \textbf{63.54} & \textbf{11.95} & \textbf{49.87} & \textbf{9.41} & \textbf{56.88} & \textbf{5.59} & \textbf{64.19} & \textbf{12.15} & \textbf{54.83} & \textbf{9.48} & \textbf{59.68} \\
\hline
\hline
\end{tabular}
}
\label{tab:main_results}
\vskip -1em
\end{table*}

Table~\ref{tab:main_results} presents the results of combining block-wise randomized Hadamard rotations with GPTQ under the MXFP4 format. Compared to the global rotation in QuaRot$^+$, BRQ delivers substantial improvements, surpassing the strong BINT4 baseline in all cases except LLaMA-3.2 3B. Particularly, on the more challenging LLaMA-3.2 1B and Qwen2.5 1.5B models, BRQ reduces perplexity from 12.78/12.80 to 11.95/12.15 and raises downstream task accuracy from 48.83/53.50 to 49.87/54.83. These results confirm that block-wise rotation is key to reconciling rotation-based methods with MXFP4, further corroborating our analysis in Section~\ref{sec:block_rotation}.

\begin{table*}[ht]
\centering
\caption{Performance comparison of optimized block rotation transformation (BRQ$_{Spin}$), random block rotation transformation (BRQ), and optimized global rotation transformation (SpinQuant).}
\resizebox{\textwidth}{!}{
\begin{tabular}{c|cc|cc|cc|cc|cc|cc}
\hline
\hline
\multirow{2}{*}{Method} & \multicolumn{2}{c|}{LLaMA 3 8B} & \multicolumn{2}{c|}{LLaMA 3.2 1B} & \multicolumn{2}{c|}{LLaMA 3.2 3B} & \multicolumn{2}{c|}{Mistral 7B} & \multicolumn{2}{c|}{Qwen2.5 1.5B} & \multicolumn{2}{c}{Qwen2.5 3B} \\
 & Wiki & Avg & Wiki & Avg & Wiki & Avg & Wiki & Avg & Wiki & Avg & Wiki & Avg \\
\hline
FP16 & 6.14 & 65.85 & 9.75 & 53.81 & 7.81 & 61.60 & 5.25 & 66.73 & 9.87 & 58.83 & 8.03 & 61.91 \\
\hline
SpinQuant & 7.62 & 61.93 & 12.72 & 49.09 & 9.85 & 56.19 & 5.68 & 63.79 & 12.64 & 53.57 & 9.58 & 59.09 \\
BRQ & 7.14 & \textbf{63.54} & 11.95 & 49.87 & 9.41 & 56.88 & 5.59 & 64.19 & 12.15 & 54.17 & 9.48 & \textbf{59.68} \\
BRQ$_{Spin}$ & \textbf{7.13} & 63.39 & \textbf{11.93} & \textbf{50.00} & \textbf{9.08} & \textbf{57.29} & \textbf{5.57} & \textbf{64.26} & \textbf{11.95} & \textbf{55.07} & \textbf{9.46} & 59.65 \\
\hline
\hline
\end{tabular}
}
\label{tab:opt_results}
\vskip -1em
\end{table*}

We next evaluate the benefits of optimizing block rotation matrices. Table~\ref{tab:opt_results} compares SpinQuant, BRQ with randomized block rotations, and BRQ$_{Spin}$, where block rotations are optimized within the SpinQuant framework. Following the SpinQuant setup, we adopt Cayley Adam~\citep{li2020efficient} and optimize using 800 sequences of length 2048 from Wikitext2.

Even without optimization, BRQ already outperforms SpinQuant with optimized global rotations. For instance, on LLaMA-3 8B, BRQ reduces perplexity from 7.68 (SpinQuant) to 7.14, also surpassing BINT4 (7.40). On downstream tasks, accuracy improves from 61.93 (SpinQuant) to 63.54, narrowing the error gap by 41\%. Similar trends hold across other models. Given that SpinQuant incurs high optimization costs~\citep{liu2024spinquant}, BRQ’s ability to achieve better performance without optimization undoubtedly enhances its practicality for real-world deployment.

Moreover, optimization brings further gains: for LLaMA-3.2 3B, perplexity decreases from 9.41 with random block rotations to 9.08 after optimization, significantly better than SpinQuant’s 9.85. However, the improvements remain limited for many models, such as LLaMA-3 8B and Mistral 7B. These findings not only demonstrate the compatibility of BRQ with existing frameworks but also suggest the potential limitations of the SpinQuant optimization scheme.

\subsection{Effect of Rotation Dimension}

\begin{wrapfigure}{r}{0.5\linewidth}
    \vskip -2em
    \centering
    \includegraphics[width=0.9\linewidth]{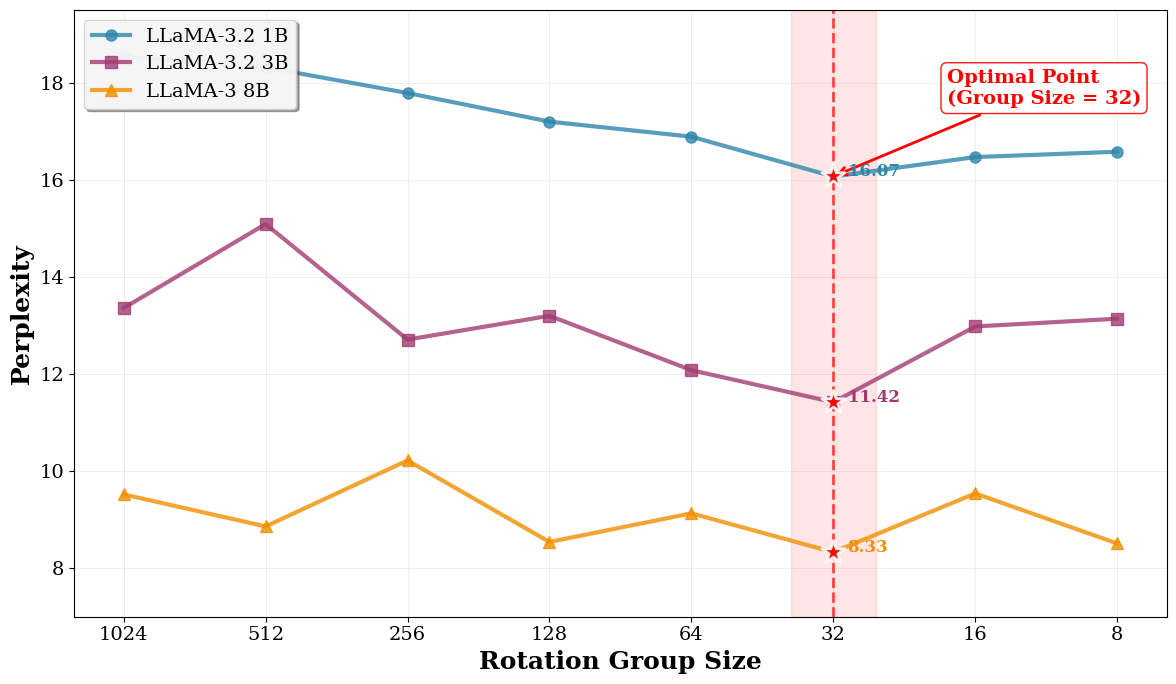}
    \caption{The effect of rotation matrix dimension on quantization accuracy.}
    \label{fig:rotation_block}
    \vskip -2em
\end{wrapfigure}

To further verify the fit of rotation dimension size to MXFP4, Figure \ref{fig:rotation_block} reports the PPL results of LLaMA-3 8B and LLaMA-3.2 1B/3B with different rotation dimensions. We observe that using larger rotation dimensions amplifies the impact of outliers and thus increases the overall loss, while smaller dimensions suffer from insufficient discrete channels. The best PPL is achieved when the rotation dimension matches the MXFP4 block size, which is consistent with our analysis in Section \ref{sec:block_rotation}.

\subsection{Performance Analysis}

\begin{table*}[!t]
\centering
\vskip -1em
\caption{Prefill latency (ms) for LLaMA-2 7B with different sequence lengths and batch sizes. Overhead is calculated relative to MXFP4.}
\setlength{\tabcolsep}{3pt} 
\small
\begin{tabular}{c|c|cc|cc|cc|cc}
\hline
\hline
\multirow{2}{*}{Batch} & \multirow{2}{*}{Method} & \multicolumn{2}{c|}{SeqLen 512} & \multicolumn{2}{c|}{SeqLen 1024} & \multicolumn{2}{c|}{SeqLen 2048} & \multicolumn{2}{c}{SeqLen 4096} \\
 & & Latency & Overhead & Latency & Overhead & Latency & Overhead & Latency & Overhead \\
\hline
  & MXFP4 & 243.09 & - & 339.31 & - & 584.31 & - & 1274.84 & - \\
1 & QuaRot & 260.18 & 7.03\% & 362.91 & 6.96\% & 618.43 & 5.84\% & 1334.03 & 4.64\% \\
  & BRQ & 253.35 & 4.22\% & 353.30 & 4.12\% & 604.25 & 3.41\% & 1307.10 & 2.53\% \\
\hline
  & MXFP4 & 732.34 & - & 1444.37 & - & 3449.58 & - & 8699.71 & - \\
8 & QuaRot & 789.13 & 7.75\% & 1547.36 & 7.13\% & 3646.83 & 5.72\% & 9032.39 & 3.82\% \\
  & BRQ & 763.72 & 4.28\% & 1499.30 & 3.80\% & 3554.16 & 3.03\% & 8816.30 & 1.34\% \\
\hline
\hline
\end{tabular}
\label{tab:latency_overhead}
\vskip -2em
\end{table*}

We implement our method on PyTorch with CUDA 12.4 and employ \textit{microsoft/microxcaling}~\citep{microxcaling} for MXFP4 quantization. In this section, we compare the performance of BRQ and QuaRot on both the prefill and decode stages using NVIDIA A800 GPUs.

We evaluate the prefill speed of LLaMA-2 7B models, with results reported in Table~\ref{tab:latency_overhead}. As expected, the block-wise rotation in BRQ drastically reduces the online rotation computation. Compared to QuaRot, BRQ lowers the additional inference latency caused by rotation by 40\%, further improving the practicality of rotation-based quantization methods. Note that since the A800 does not natively support FP4 inference, inference speed would be further improved on dedicated MXFP4 hardware.

\section{Conclusion and Future Work}
In this study, we benchmarked representative INT4 PTQ algorithms under the MXFP4 format, systematically revealing the limitations of existing methods in this hardware-friendly setting. We analyzed the incompatibility between MXFP4 and rotation-based approaches, identifying the conflict between block-wise quantization and rotation-induced energy redistribution, amplified by MXFP4’s PoT scale. To address this, we introduce BRQ, which adapts rotation to MXFP4, resolving compatibility issues and significantly improving PTQ performance.

Looking ahead, we plan to explore better rotation matrix optimization solutions, and whether replacing the online fast Hadamard transform ($R_4$) with optimized block-wise rotations can further enhance accuracy while carefully balancing inference latency. This work aims to provide both theoretical insights and practical guidance for deploying large language models on next-generation low-bit floating-point hardware.






\begin{thebibliography}{38}
\providecommand{\natexlab}[1]{#1}
\providecommand{\url}[1]{\texttt{#1}}
\expandafter\ifx\csname urlstyle\endcsname\relax
  \providecommand{\doi}[1]{doi: #1}\else
  \providecommand{\doi}{doi: \begingroup \urlstyle{rm}\Url}\fi

\bibitem[AMD(2025)]{amd_quark}
AMD.
\newblock Amd quark model optimizer, 2025.
\newblock URL \url{https://github.com/amd/Quark}.
\newblock Accessed: 2025-09-02.

\bibitem[Ashkboos et~al.(2024)Ashkboos, Mohtashami, Croci, Li, Cameron, Jaggi, Alistarh, Hoefler, and Hensman]{ashkboos2024quarot}
Saleh Ashkboos, Amirkeivan Mohtashami, Maximilian~L Croci, Bo~Li, Pashmina Cameron, Martin Jaggi, Dan Alistarh, Torsten Hoefler, and James Hensman.
\newblock Quarot: Outlier-free 4-bit inference in rotated llms.
\newblock \emph{Advances in Neural Information Processing Systems}, 37:\penalty0 100213--100240, 2024.

\bibitem[Behdin et~al.(2023)Behdin, Acharya, Gupta, Song, Zhu, Keerthi, and Mazumder]{behdin2023quantease}
Kayhan Behdin, Ayan Acharya, Aman Gupta, Qingquan Song, Siyu Zhu, Sathiya Keerthi, and Rahul Mazumder.
\newblock Quantease: Optimization-based quantization for language models.
\newblock \emph{arXiv preprint arXiv:2309.01885}, 2023.

\bibitem[Bisk et~al.(2020)Bisk, Zellers, Gao, Choi, et~al.]{bisk2020piqa}
Yonatan Bisk, Rowan Zellers, Jianfeng Gao, Yejin Choi, et~al.
\newblock Piqa: Reasoning about physical commonsense in natural language.
\newblock In \emph{Proceedings of the AAAI conference on artificial intelligence}, volume~34, pp.\  7432--7439, 2020.

\bibitem[Boratko et~al.(2018)Boratko, Padigela, Mikkilineni, Yuvraj, Das, McCallum, Chang, Fokoue-Nkoutche, Kapanipathi, Mattei, et~al.]{boratko2018systematic}
Michael Boratko, Harshit Padigela, Divyendra Mikkilineni, Pritish Yuvraj, Rajarshi Das, Andrew McCallum, Maria Chang, Achille Fokoue-Nkoutche, Pavan Kapanipathi, Nicholas Mattei, et~al.
\newblock A systematic classification of knowledge, reasoning, and context within the arc dataset.
\newblock \emph{arXiv preprint arXiv:1806.00358}, 2018.

\bibitem[Chee et~al.(2024)Chee, Cai, Kuleshov, and De~Sa]{chee2024quip}
Jerry Chee, Yaohui Cai, Volodymyr Kuleshov, and Christopher~M De~Sa.
\newblock Quip: 2-bit quantization of large language models with guarantees.
\newblock \emph{Advances in Neural Information Processing Systems}, 36, 2024.

\bibitem[Czak{\'o} et~al.(2025)Czak{\'o}, Kert{\'e}sz, and Sz{\'e}n{\'a}si]{czako2025addressing}
Patrik Czak{\'o}, G{\'a}bor Kert{\'e}sz, and S{\'a}ndor Sz{\'e}n{\'a}si.
\newblock Addressing activation outliers in llms: A systematic review of post-training quantization techniques.
\newblock \emph{IEEE Access}, 2025.

\bibitem[Dantas et~al.(2025)Dantas, Cordeiro, and Junior]{dantas2025review}
Pierre~V Dantas, Lucas~C Cordeiro, and Waldir~SS Junior.
\newblock A review of state-of-the-art techniques for large language model compression.
\newblock \emph{Complex \& Intelligent Systems}, 11\penalty0 (9):\penalty0 1--40, 2025.

\bibitem[Darvish~Rouhani et~al.(2023)Darvish~Rouhani, Zhao, Elango, Shafipour, Hall, Mesmakhosroshahi, More, Melnick, Golub, Varatkar, et~al.]{darvish2023shared}
Bita Darvish~Rouhani, Ritchie Zhao, Venmugil Elango, Rasoul Shafipour, Mathew Hall, Maral Mesmakhosroshahi, Ankit More, Levi Melnick, Maximilian Golub, Girish Varatkar, et~al.
\newblock With shared microexponents, a little shifting goes a long way.
\newblock In \emph{Proceedings of the 50th Annual International Symposium on Computer Architecture}, pp.\  1--13, 2023.

\bibitem[Dettmers et~al.(2022)Dettmers, Lewis, Belkada, and Zettlemoyer]{dettmers2022gpt3}
Tim Dettmers, Mike Lewis, Younes Belkada, and Luke Zettlemoyer.
\newblock Gpt3. int8 (): 8-bit matrix multiplication for transformers at scale.
\newblock \emph{Advances in neural information processing systems}, 35:\penalty0 30318--30332, 2022.

\bibitem[Elangovan et~al.(2025)Elangovan, Sakr, Raghunathan, and Khailany]{elangovan2025bcq}
Reena Elangovan, Charbel Sakr, Anand Raghunathan, and Brucek Khailany.
\newblock Bcq: Block clustered quantization for 4-bit (w4a4) llm inference.
\newblock \emph{arXiv preprint arXiv:2502.05376}, 2025.

\bibitem[Frantar et~al.(2022)Frantar, Ashkboos, Hoefler, and Alistarh]{frantar2022gptq}
Elias Frantar, Saleh Ashkboos, Torsten Hoefler, and Dan Alistarh.
\newblock Gptq: Accurate post-training quantization for generative pre-trained transformers.
\newblock \emph{arXiv preprint arXiv:2210.17323}, 2022.

\bibitem[Guan et~al.(2024)Guan, Huang, Su, Huang, Wong, and Yu]{guan2024aptq}
Ziyi Guan, Hantao Huang, Yupeng Su, Hong Huang, Ngai Wong, and Hao Yu.
\newblock Aptq: Attention-aware post-training mixed-precision quantization for large language models.
\newblock In \emph{Proceedings of the 61st ACM/IEEE Design Automation Conference}, pp.\  1--6, 2024.

\bibitem[Han et~al.(2025)Han, Cheng, Wang, Lu, Wang, Xu, Yang, Jiang, et~al.]{han2025bbal}
Xiaomeng Han, Yuan Cheng, Jing Wang, Junyang Lu, Hui Wang, Ning Xu, Dawei Yang, Zhe Jiang, et~al.
\newblock Bbal: A bidirectional block floating point-based quantisation accelerator for large language models.
\newblock \emph{arXiv preprint arXiv:2504.15721}, 2025.

\bibitem[Kim et~al.(2024)Kim, Kim, Cho, Lee, Kim, and Jeon]{kim2024boa}
Junhan Kim, Ho-young Kim, Eulrang Cho, Chungman Lee, Joonyoung Kim, and Yongkweon Jeon.
\newblock Boa: Attention-aware post-training quantization without backpropagation.
\newblock \emph{arXiv preprint arXiv:2406.13474}, 2024.

\bibitem[Lee et~al.(2024)Lee, Park, Kim, Kim, Oh, Oh, and Choi]{lee2024amxfp4}
Janghwan Lee, Jiwoong Park, Jinseok Kim, Yongjik Kim, Jungju Oh, Jinwook Oh, and Jungwook Choi.
\newblock Amxfp4: Taming activation outliers with asymmetric microscaling floating-point for 4-bit llm inference.
\newblock \emph{arXiv preprint arXiv:2411.09909}, 2024.

\bibitem[Li et~al.(2020)Li, Fuxin, and Todorovic]{li2020efficient}
Jun Li, Li~Fuxin, and Sinisa Todorovic.
\newblock Efficient riemannian optimization on the stiefel manifold via the cayley transform.
\newblock \emph{arXiv preprint arXiv:2002.01113}, 2020.

\bibitem[Lin et~al.(2024{\natexlab{a}})Lin, Xu, Wu, Cui, Zhang, Mou, Song, Sun, and Wei]{lin2024duquant}
Haokun Lin, Haobo Xu, Yichen Wu, Jingzhi Cui, Yingtao Zhang, Linzhan Mou, Linqi Song, Zhenan Sun, and Ying Wei.
\newblock Duquant: Distributing outliers via dual transformation makes stronger quantized llms.
\newblock \emph{Advances in Neural Information Processing Systems}, 37:\penalty0 87766--87800, 2024{\natexlab{a}}.

\bibitem[Lin et~al.(2024{\natexlab{b}})Lin, Tang, Yang, Zhang, Xiao, Gan, and Han]{lin2024qserve}
Yujun Lin, Haotian Tang, Shang Yang, Zhekai Zhang, Guangxuan Xiao, Chuang Gan, and Song Han.
\newblock Qserve: W4a8kv4 quantization and system co-design for efficient llm serving.
\newblock \emph{arXiv preprint arXiv:2405.04532}, 2024{\natexlab{b}}.

\bibitem[Liu et~al.(2024{\natexlab{a}})Liu, Wen, Wang, Ye, Zhang, Cao, Li, and Yang]{liu2024vptq}
Yifei Liu, Jicheng Wen, Yang Wang, Shengyu Ye, Li~Lyna Zhang, Ting Cao, Cheng Li, and Mao Yang.
\newblock Vptq: Extreme low-bit vector post-training quantization for large language models.
\newblock \emph{arXiv preprint arXiv:2409.17066}, 2024{\natexlab{a}}.

\bibitem[Liu et~al.(2024{\natexlab{b}})Liu, Zhao, Fedorov, Soran, Choudhary, Krishnamoorthi, Chandra, Tian, and Blankevoort]{liu2024spinquant}
Zechun Liu, Changsheng Zhao, Igor Fedorov, Bilge Soran, Dhruv Choudhary, Raghuraman Krishnamoorthi, Vikas Chandra, Yuandong Tian, and Tijmen Blankevoort.
\newblock Spinquant: Llm quantization with learned rotations.
\newblock \emph{arXiv preprint arXiv:2405.16406}, 2024{\natexlab{b}}.

\bibitem[Luo et~al.(2025)Luo, Pandey, Yang, Ghodsi, Yang, Burbank, Wang, Gogineni, Wang, Gu, Fu, Qin, and Gupta]{amd2025gptoss}
Andy Luo, Shekhar Pandey, Hongxia Yang, Mahdi Ghodsi, Charles Yang, Niles Burbank, George Wang, Kailash Gogineni, Xun Wang, Zhenyu Gu, Yao Fu, Yanyuan Qin, and Anshul Gupta.
\newblock How to run openai’s gpt-oss 20b and 120b models on amd ryzen™ ai processors and radeon™ graphics cards, August 2025.
\newblock URL \url{https://www.amd.com/en/blogs/2025/how-to-run-openai-gpt-oss-20b-120b-models-on-amd-ryzen-ai-radeon.html}.
\newblock Accessed: 2025-09-16.

\bibitem[Ma et~al.(2024)Ma, Li, Zheng, Ling, Xiao, Wang, Wen, Chao, and Ji]{ma2024affinequant}
Yuexiao Ma, Huixia Li, Xiawu Zheng, Feng Ling, Xuefeng Xiao, Rui Wang, Shilei Wen, Fei Chao, and Rongrong Ji.
\newblock Affinequant: Affine transformation quantization for large language models.
\newblock \emph{arXiv preprint arXiv:2403.12544}, 2024.

\bibitem[Microsoft(2024)]{microxcaling}
Microsoft.
\newblock microxcaling.
\newblock \url{https://github.com/microsoft/microxcaling}, 2024.
\newblock Accessed: 2025-08-26.

\bibitem[Mihaylov et~al.(2018)Mihaylov, Clark, Khot, and Sabharwal]{mihaylov2018can}
Todor Mihaylov, Peter Clark, Tushar Khot, and Ashish Sabharwal.
\newblock Can a suit of armor conduct electricity? a new dataset for open book question answering.
\newblock \emph{arXiv preprint arXiv:1809.02789}, 2018.

\bibitem[NVIDIA(2023)]{tensorrt_llm}
NVIDIA.
\newblock Tensorrt-llm: Open-source library for optimizing large language model inference, 2023.
\newblock URL \url{https://github.com/NVIDIA/TensorRT-LLM}.
\newblock Accessed: 2025-08-20.

\bibitem[OpenAI(2025)]{openai_gpt_oss}
OpenAI.
\newblock Introducing gpt-oss, 2025.
\newblock URL \url{https://openai.com/index/introducing-gpt-oss/}.
\newblock Accessed: 2025-09-02.

\bibitem[Rouhani et~al.(2023)Rouhani, Garegrat, Savell, More, Han, Zhao, Hall, Klar, Chung, Yu, Schulte, Wittig, Bratt, Stephens, Milanovic, Brothers, Dubey, Cornea, Heinecke, Rodriguez, Langhammer, Deng, Naumov, Micikevicius, Siu, and Verrilli]{ocp_mx_spec_2023}
Bita~Darvish Rouhani, Nitin Garegrat, Tom Savell, Ankit More, Kyung-Nam Han, Ritchie Zhao, Mathew Hall, Jasmine Klar, Eric Chung, Yuan Yu, Michael Schulte, Ralph Wittig, Ian Bratt, Nigel Stephens, Jelena Milanovic, John Brothers, Pradeep Dubey, Marius Cornea, Alexander Heinecke, Andres Rodriguez, Martin Langhammer, Summer Deng, Maxim Naumov, Paulius Micikevicius, Michael Siu, and Colin Verrilli.
\newblock Ocp microscaling formats (mx) specification version 1.0, 2023.
\newblock URL \url{https://www.opencompute.org/documents/ocp-microscaling-formats-mx-v1-0-spec-final-pdf}.
\newblock Accessed: 2025-08-20.

\bibitem[Sadegh~Akhondzadeh et~al.(2025)Sadegh~Akhondzadeh, Bojchevski, Eleftheriou, and Dazzi]{sadegh2025kurtail}
Mohammad Sadegh~Akhondzadeh, Aleksandar Bojchevski, Evangelos Eleftheriou, and Martino Dazzi.
\newblock Kurtail: Kurtosis-based llm quantization.
\newblock \emph{arXiv e-prints}, pp.\  arXiv--2503, 2025.

\bibitem[Sakaguchi et~al.(2021)Sakaguchi, Bras, Bhagavatula, and Choi]{sakaguchi2021winogrande}
Keisuke Sakaguchi, Ronan~Le Bras, Chandra Bhagavatula, and Yejin Choi.
\newblock Winogrande: An adversarial winograd schema challenge at scale.
\newblock \emph{Communications of the ACM}, 64\penalty0 (9):\penalty0 99--106, 2021.

\bibitem[Shao et~al.(2023)Shao, Chen, Zhang, Xu, Zhao, Li, Zhang, Gao, Qiao, and Luo]{shao2023omniquant}
Wenqi Shao, Mengzhao Chen, Zhaoyang Zhang, Peng Xu, Lirui Zhao, Zhiqian Li, Kaipeng Zhang, Peng Gao, Yu~Qiao, and Ping Luo.
\newblock Omniquant: Omnidirectionally calibrated quantization for large language models.
\newblock \emph{arXiv preprint arXiv:2308.13137}, 2023.

\bibitem[Sun et~al.(2024)Sun, Liu, Bai, Bao, Zhao, Li, Hu, Yu, Hou, Yuan, et~al.]{sun2024flatquant}
Yuxuan Sun, Ruikang Liu, Haoli Bai, Han Bao, Kang Zhao, Yuening Li, Jiaxin Hu, Xianzhi Yu, Lu~Hou, Chun Yuan, et~al.
\newblock Flatquant: Flatness matters for llm quantization.
\newblock \emph{arXiv preprint arXiv:2410.09426}, 2024.

\bibitem[Sung et~al.(2025)Sung, Yadav, Li, Yoon, and Bansal]{sung2025rsq}
Yi-Lin Sung, Prateek Yadav, Jialu Li, Jaehong Yoon, and Mohit Bansal.
\newblock Rsq: Learning from important tokens leads to better quantized llms.
\newblock \emph{arXiv preprint arXiv:2503.01820}, 2025.

\bibitem[Team(2024)]{team2024qwen2}
Qwen Team.
\newblock Qwen2 technical report.
\newblock \emph{arXiv preprint arXiv:2407.10671}, 2024.

\bibitem[Tseng et~al.(2024)Tseng, Chee, Sun, Kuleshov, and De~Sa]{tseng2024quip}
Albert Tseng, Jerry Chee, Qingyao Sun, Volodymyr Kuleshov, and Christopher De~Sa.
\newblock Quip\#: Even better llm quantization with hadamard incoherence and lattice codebooks.
\newblock \emph{arXiv preprint arXiv:2402.04396}, 2024.

\bibitem[van Breugel et~al.(2025)van Breugel, Bondarenko, Whatmough, and Nagel]{van2025fptquant}
Boris van Breugel, Yelysei Bondarenko, Paul Whatmough, and Markus Nagel.
\newblock Fptquant: Function-preserving transforms for llm quantization.
\newblock \emph{arXiv preprint arXiv:2506.04985}, 2025.

\bibitem[Wu et~al.(2025)Wu, Yang, Zhan, Yuan, Chao, and Wong]{wu2025survey}
Junchao Wu, Shu Yang, Runzhe Zhan, Yulin Yuan, Lidia~Sam Chao, and Derek~Fai Wong.
\newblock A survey on llm-generated text detection: Necessity, methods, and future directions.
\newblock \emph{Computational Linguistics}, 51\penalty0 (1):\penalty0 275--338, 2025.

\bibitem[Xiao et~al.(2023)Xiao, Lin, Seznec, Wu, Demouth, and Han]{xiao2023smoothquant}
Guangxuan Xiao, Ji~Lin, Mickael Seznec, Hao Wu, Julien Demouth, and Song Han.
\newblock Smoothquant: Accurate and efficient post-training quantization for large language models.
\newblock In \emph{International Conference on Machine Learning}, pp.\  38087--38099. PMLR, 2023.

\end{thebibliography}
\bibliographystyle{iclr2026_conference}


\newpage
\appendix
\section{Appendix}
\subsection{Use of LLMs}
The text in this paper has been professionally refined using LLM to enhance clarity, coherence, and adherence to academic writing standards.

\subsection{Experimental Details} \label{apx:setting_details}
In this section, we provide the detailed hyperparameter settings used for each method in the benchmark, to ensure reproducibility of our experiments. All methods employ WikiText2 as the calibration dataset, and were conducted on NVIDIA A800 GPUs.

\begin{itemize}
\item \textbf{GPTQ:} Calibrated with 128 sequences of length 2048. The damping parameter for Hessian estimation is set to 0.01, following the authors’ recommendation~\citep{frantar2022gptq}.

\item \textbf{SmoothQuant:} Calibrated with 512 sequences of length 512 for activation scaling. The smoothing coefficient $\alpha$ is set to 0.85, consistent with the official repository~\citep{xiao2023smoothquant}.

\item \textbf{OmniQuant:} Calibrated with 128 sequences of length 2048, with optimization applied to learnable weight clipping and equivalent transformations~\citep{shao2023omniquant}.

\item \textbf{SpinQuant:} Calibrated with 800 sequences of length 2048, optimized via the Cayley SGD~\citep{li2020efficient} optimizer for rotation matrix training~\citep{liu2024spinquant}.
\end{itemize}

\subsection{Application on 70B model} \label{apx:vs70b}

To explore the effect of BRQ on larger-scale models, we conducted comparative experiments on LLaMA-2 70B. As shown in Table~\ref{tab:llama2-70b-results}, BRQ consistently outperforms existing methods at this scale. In particular, compared with QuaRot—which also leverages randomized Hadamard transforms—BRQ reduces the perplexity of LLaMA-2 70B from 3.76 to 3.62, while improving the average downstream accuracy from 68.86 to 69.10. Due to memory limitations on our available servers, we were unable to apply SpinQuant to the 70B model, and thus its results are not reported here. These results further demonstrate the effectiveness of BRQ on large-scale models.

\begin{table}[H]
\centering
\caption{Performance of different quantization methods on LLaMA-2 70B.}
\begin{tabular}{c|c|ccccc|c}
\hline
\hline
Method & WiKi & WG & PIQA & OBQA & ARC-E & ARC-C & Avg \\
\hline
FP16 & 3.32 & 77.97 & 82.75 & 48.80 & 81.01 & 57.50 & 69.61 \\
RTN    & 4.20 & 75.92 & 80.90 & 46.00 & 78.53 & 53.32 & 66.93 \\
GPTQ     & 3.79 & 75.37 & 80.84 & 48.00 & 79.92 & 54.60 & 67.75 \\
QuaRot$^+$  & 3.76 & 77.66 & 81.82 & 47.80 & 79.75 & 57.25 & 68.86 \\
BRQ      & 3.62 & 77.68 & 82.54 & 47.60 & 80.30 & 57.40 & 69.10 \\
\hline
\hline
\end{tabular}
\label{tab:llama2-70b-results}
\end{table}

\subsection{Rotation Comparison} \label{apx:rot}
\begin{figure*}[ht]
\begin{center}
\centerline{\includegraphics[width=\textwidth]{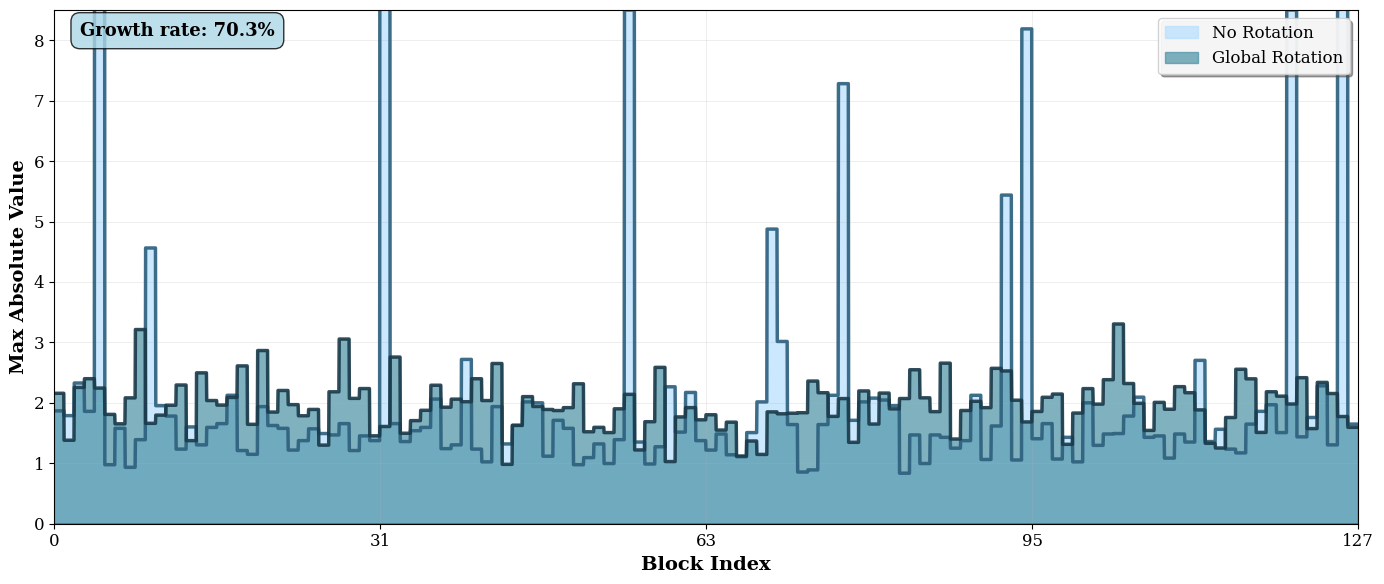}}
\caption{Changes in block maximum values after applying global rotation.}
\label{fig:no vs. gr}
\end{center}
\vskip -2em
\end{figure*}

Figure~\ref{fig:no vs. gr} illustrates the block-wise scale distribution (defined as the maximum absolute value within each block) under the original activations and after applying global rotation. In the original activations, only a small fraction of blocks contain outliers, while most blocks maintain relatively small scales. After global rotation, although the prominent outliers are mitigated, the scales of more than 70\% of the regular blocks increase substantially. This amplification of regular-block scales is the fundamental reason behind the collapse of quantization performance under MXFP4 following global rotation.

\begin{figure*}[ht]
\begin{center}
\centerline{\includegraphics[width=\textwidth]{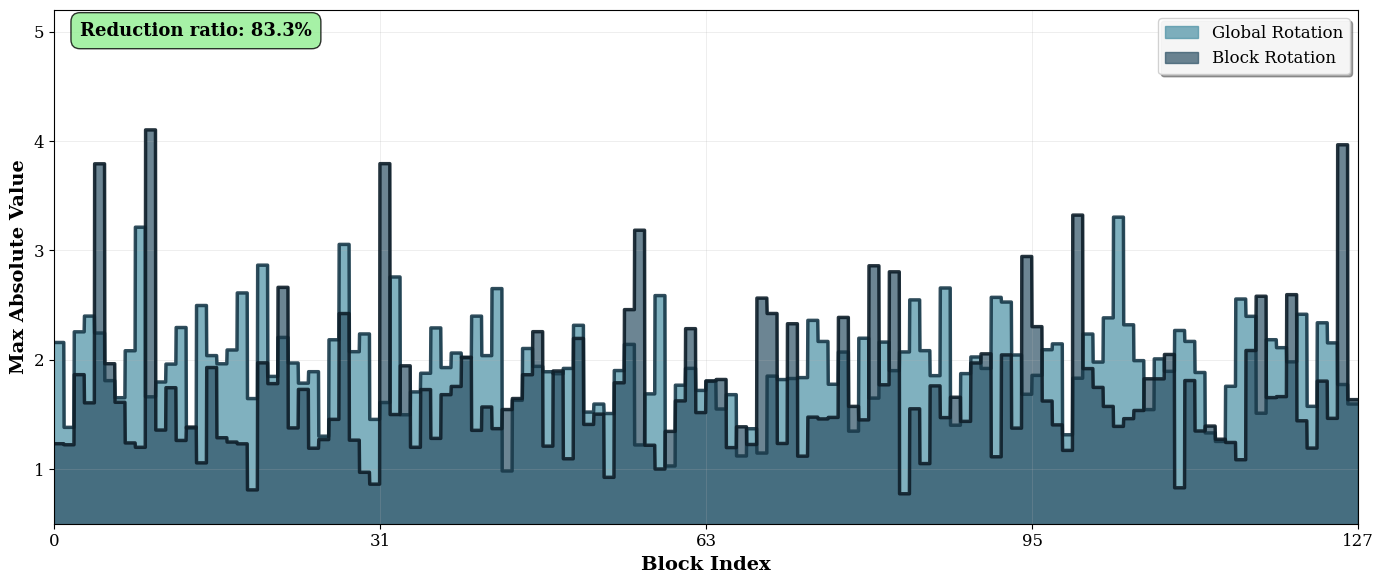}}
\caption{Comparison of group maximum values after global rotation and block rotation.}
\label{fig:br vs. gr}
\end{center}
\vskip -2em
\end{figure*}

We further analyze the blocks that exhibited significant scale growth in Figure~\ref{fig:br vs. gr} and evaluate the impact of block-wise rotation on these blocks. Figure 8 compares the scale distributions under global rotation and block-wise rotation. We observe that block-wise rotation mitigates over 80\% of the scale inflation introduced by global rotation. This result provides strong evidence that block-wise rotation effectively alleviates the quantization degradation caused by global rotation, thereby confirming its effectiveness.

\subsection{Detailed Results} \label{apx:results}

This section presents the detailed results of the experiments reported in the main text. Notably, in these experiments, BRQ employs block-wise stochastic Hadamard matrices for rotation without any additional optimization of the rotation matrices.

\subsubsection{LLaMA Results}
The following are the detailed experimental results of the LLaMA family series models in Tables~\ref{tab:benchmark} and~\ref{tab:main_results}.

\begin{table}[H]
\centering
\caption{Evaluation of different methods on LLaMA-2 7B across multiple benchmarks.}
\begin{tabular}{cccccccc}
\hline
\hline
\textbf{Method} & \textbf{WiKi} & \textbf{WG} & \textbf{PIQA} & \textbf{OBQA} & \textbf{ARC-E} & \textbf{ARC-C} & \textbf{Avg} \\
\hline
Baseline & 5.47  & 68.98 & 79.05 & 44.20 & 74.57 & 46.16 & 62.59 \\
BINT4 & 5.94 & 68.19 & 76.93 & 43.00 & 73.65 & 44.71 & 61.30 \\
\hline
RTN            & 7.08  & 64.80 & 76.39 & 39.20 & 65.70 & 40.19 & 57.26 \\
SmoothQuant    & 7.04  & 64.64 & 76.17 & 39.00 & 66.75 & 39.33 & 57.18 \\
GPTQ           & 6.56  & 66.61 & 76.55 & 40.60 & 71.12 & 41.46 & 59.27 \\
OmniQuant      & 6.56  & 63.06 & 76.33 & 36.60 & 67.09 & 40.27 & 56.67 \\
QuaRot         & 13.09 & 59.11 & 71.21 & 34.40 & 55.05 & 31.82 & 50.32 \\
QuaRot$^+$        & 6.29  & 67.24 & 75.68 & 39.80 & 69.02 & 40.01 & 58.35 \\
SpinQuant      & 5.99  & 66.14 & 77.69 & 40.40 & 70.58 & 41.38 & 59.24 \\
BRQ & 5.84 & 67.09 & 76.77 & 44.80 & 73.23 & 43.17 & 61.01 \\
\hline
\hline
\end{tabular}
\label{tab:llama2_7b_quant}
\end{table}

\begin{table}[H]
\centering
\caption{Evaluation of different methods on LLaMA-3 8B across multiple benchmarks.}
\begin{tabular}{cccccccc}
\hline
\hline
\textbf{Method} & \textbf{WiKi} & \textbf{WG} & \textbf{PIQA} & \textbf{OBQA} & \textbf{ARC-E} & \textbf{ARC-C} & \textbf{Avg} \\
\hline
Baseline       & 6.14  & 73.16 & 80.57 & 44.80 & 77.56 & 53.15 & 65.85 \\
BINT4 & 7.40 & 70.40 & 78.94 & 43.60 & 73.95 & 48.72 & 63.12 \\
\hline
RTN            & 8.23  & 67.56 & 77.31 & 41.80 & 70.74 & 45.64 & 60.61 \\
SmoothQuant    & 8.11  & 67.88 & 78.29 & 43.60 & 71.00 & 45.31 & 61.22 \\
GPTQ           & 7.68  & 70.48 & 76.06 & 42.00 & 73.23 & 45.64 & 61.48 \\
OmniQuant      & 8.16  & 66.06 & 77.20 & 40.60 & 72.85 & 45.65 & 60.47 \\
QuaRot         & 9.56  & 67.71 & 75.57 & 41.60 & 69.02 & 42.40 & 59.26 \\
QuaRot$^+$        & 7.68  & 68.16 & 75.36 & 42.80 & 72.66 & 48.89 & 61.57 \\
SpinQuant      & 7.62  & 69.56 & 76.93 & 42.00 & 72.90 & 48.25 & 61.93 \\
BRQ & 7.14 & 71.98 & 78.51 & 42.60 & 75.04 & 49.57 & 63.54 \\
\hline
\hline
\end{tabular}
\label{tab:llama3_8b_quant}
\end{table}

\begin{table}[H]
\centering
\caption{Evaluation of different methods on LLaMA-2 13B across multiple benchmarks.}
\begin{tabular}{cccccccc}
\hline
\hline
\textbf{Method} & \textbf{WiKi} & \textbf{WG} & \textbf{PIQA} & \textbf{OBQA} & \textbf{ARC-E} & \textbf{ARC-C} & \textbf{Avg} \\
\hline
Baseline       & 4.88  & 72.13 & 80.52 & 45.20 & 77.44 & 49.14 & 64.89 \\
BINT4 & 5.16 & 72.06 & 78.84 & 43.20 & 75.55 & 46.93 & 63.32 \\ 
\hline
RTN            & 5.90  & 69.45 & 77.25 & 42.60 & 72.64 & 45.05 & 61.40 \\
SmoothQuant    & 5.73  & 69.14 & 77.37 & 42.40 & 72.98 & 45.73 & 61.52 \\
GPTQ           & 5.41  & 70.56 & 78.56 & 44.40 & 75.04 & 45.98 & 62.91 \\
OmniQuant      & 5.43  & 68.43 & 78.29 & 42.00 & 74.58 & 46.16 & 61.89 \\
QuaRot         & 7.03  & 65.43 & 76.87 & 40.20 & 71.08 & 41.89 & 59.09 \\
QuaRot$^+$        & 5.57  & 67.79 & 78.18 & 40.60 & 74.62 & 46.67 & 61.57 \\
SpinQuant      & 5.20  & 68.98 & 78.45 & 42.80 & 75.63 & 48.04 & 62.78 \\
BRQ & 5.19 & 70.48 & 79.27 & 43.20 & 75.76 & 47.56 & 63.25 \\
\hline
\hline
\end{tabular}
\label{tab:llama2_13b_quant}
\end{table}

\begin{table}[H]
\centering
\caption{Evaluation of different methods on LLaMA-3.2 1B across multiple benchmarks.}
\begin{tabular}{cccccccc}
\hline
\hline
\textbf{Method} & \textbf{WiKi} & \textbf{WG} & \textbf{PIQA} & \textbf{OBQA} & \textbf{ARC-E} & \textbf{ARC-C} & \textbf{Avg} \\
\hline
Baseline       & 9.75  & 60.61 & 74.53 & 37.20 & 60.47 & 36.26 & 53.81 \\
BINT4 & 13.56 & 54.78 & 69.26 & 34.80 & 52.74 & 30.20 & 48.36 \\
\hline
RTN            & 15.91 & 54.30 & 66.10 & 32.80 & 50.37 & 30.88 & 46.89 \\
SmoothQuant    & 16.86 & 55.72 & 66.27 & 30.60 & 50.55 & 29.27 & 46.48 \\
GPTQ           & 13.35 & 56.66 & 69.58 & 32.40 & 52.48 & 31.48 & 48.52 \\
OmniQuant      & 14.32 & 55.09 & 68.12 & 32.80 & 53.37 & 31.48 & 48.17 \\
QuaRot         & 17.86 & 55.40 & 66.05 & 29.40 & 47.93 & 28.32 & 45.42 \\
QuaRot$^+$        & 12.78 & 56.74 & 69.85 & 32.60 & 53.42 & 31.56 & 48.83 \\
SpinQuant      & 12.72 & 55.38 & 70.67 & 33.00 & 53.87 & 32.51 & 49.09 \\
BRQ & 11.95 & 55.88 & 70.46 & 34.20 & 55.47 & 33.36 & 49.87 \\
\hline
\hline
\end{tabular}
\label{tab:llama3.2_1b_quant}
\end{table}

\begin{table}[H]
\centering
\caption{Evaluation of different methods on LLaMA-3.2 3B across multiple benchmarks.}
\begin{tabular}{cccccccc}
\hline
\hline
\textbf{Method} & \textbf{WiKi} & \textbf{WG} & \textbf{PIQA} & \textbf{OBQA} & \textbf{ARC-E} & \textbf{ARC-C} & \textbf{Avg} \\
\hline
Baseline       & 7.81  & 69.37 & 77.52 & 43.40 & 71.63 & 46.07 & 61.60 \\
BINT4 & 9.29 & 64.64 & 75.19 & 38.60 & 66.50 & 42.41 & 57.47 \\ 
\hline
RTN            & 10.27 & 63.93 & 73.06 & 40.00 & 62.16 & 36.94 & 55.22 \\
SmoothQuant    & 10.38 & 62.27 & 73.05 & 38.40 & 62.93 & 38.59 & 55.05 \\
GPTQ           & 9.50  & 63.29 & 73.55 & 37.00 & 63.04 & 40.10 & 55.40 \\
OmniQuant      & 9.85  & 61.01 & 75.84 & 38.40 & 63.30 & 40.27 & 55.76 \\
QuaRot         & 13.36 & 59.43 & 70.23 & 35.60 & 57.40 & 35.32 & 51.60 \\
QuaRot$^+$        & 9.92  & 65.43 & 73.18 & 37.80 & 64.65 & 38.48 & 55.91 \\
SpinQuant      & 9.85  & 65.11 & 73.61 & 38.60 & 64.87 & 38.74 & 56.19 \\
BRQ & 9.41 & 62.59 & 75.35 & 38.60 & 67.42 & 40.44 & 56.88 \\
\hline
\hline
\end{tabular}
\label{tab:llama3.2_3b_quant}
\end{table}

\subsubsection{Mistral Results}
The following are the detailed experimental results of the Mistral family series models in Tables~\ref{tab:benchmark} and~\ref{tab:main_results}.

\begin{table}[H]
\centering
\caption{Evaluation of different methods on Mistral 7B across multiple benchmarks.}
\begin{tabular}{cccccccc}
\hline
\hline
\textbf{Method} & \textbf{WiKi} & \textbf{WG} & \textbf{PIQA} & \textbf{OBQA} & \textbf{ARC-E} & \textbf{ARC-C} & \textbf{Avg} \\
\hline
Baseline       & 5.25  & 73.95 & 82.10 & 44.00 & 79.50 & 54.09 & 66.73 \\
BINT4 & 5.63 & 72.06 & 80.41 & 44.80 & 77.57 & 51.54 & 65.28 \\ 
\hline
RTN            & 6.56  & 69.06 & 80.73 & 43.00 & 74.66 & 46.84 & 62.86 \\
SmoothQuant    & 6.49  & 70.64 & 79.71 & 42.00 & 75.08 & 47.35 & 62.96 \\
GPTQ           & 6.00  & 69.53 & 79.48 & 43.20 & 75.71 & 48.80 & 63.34 \\
OmniQuant      & 6.37  & 67.25 & 78.56 & 40.20 & 75.13 & 47.95 & 61.82 \\
QuaRot         & 6.65  & 68.27 & 78.94 & 39.00 & 71.42 & 44.02 & 60.33 \\
QuaRot$^+$        & 5.73  & 69.61 & 80.84 & 42.80 & 76.85 & 48.20 & 63.66 \\
SpinQuant      & 5.68  & 71.90 & 79.71 & 42.80 & 76.35 & 48.21 & 63.79 \\
BRQ & 5.59 & 71.72 & 80.69 & 42.80 & 76.68 & 49.06 & 64.19 \\
\hline
\hline
\end{tabular}
\label{tab:mistral7b_quant}
\end{table}

\subsubsection{Qwen Results}
The following are the detailed experimental results of the Qwen2.5 family series models in Tables~\ref{tab:main_results}.

\begin{table}[H]
\centering
\caption{Evaluation of different methods on Qwen2.5 1.5B across multiple benchmarks.}
\begin{tabular}{cccccccc}
\hline
\hline
\textbf{Method} & \textbf{WiKi} & \textbf{WG} & \textbf{PIQA} & \textbf{OBQA} & \textbf{ARC-E} & \textbf{ARC-C} & \textbf{Avg} \\
\hline
Baseline & 9.87 & 62.98 & 75.41 & 41.20 & 71.13 & 43.43 & 58.83 \\
BINT4 & 13.98 & 58.43 & 70.44 & 37.00 & 66.67 & 37.36 & 53.98 \\
\hline
RTN & 16.61 & 57.93 & 70.51 & 36.60 & 61.20 & 37.20 & 52.69 \\
GPTQ & 13.94 & 58.14 & 70.09 & 37.40 & 62.90 & 37.14 & 53.13 \\
QuaRot & 16.33 & 56.75 & 70.29 & 32.00 & 61.03 & 36.75 & 51.36 \\
QuaRot$^+$ & 12.80 & 58.96 & 71.87 & 36.80 & 62.42 & 37.46 & 53.50 \\
SpinQuant & 12.64 & 59.91 & 71.22 & 36.60 & 62.58 & 37.52 & 53.57 \\
BRQ & 12.15 & 58.96 & 71.87 & 36.80 & 67.00 & 39.51 & 54.83 \\
\hline
\hline
\end{tabular}
\label{tab:qwen1.5b_quant}
\end{table}

\begin{table}[H]
\centering
\caption{Evaluation of different methods on Qwen2.5 3B across multiple benchmarks.}
\begin{tabular}{cccccccc}
\hline
\hline
\textbf{Method} & \textbf{WiKi} & \textbf{WG} & \textbf{PIQA} & \textbf{OBQA} & \textbf{ARC-E} & \textbf{ARC-C} & \textbf{Avg} \\
\hline
Baseline & 8.03 & 68.90 & 78.67 & 41.80 & 73.27 & 46.93 & 61.91 \\
BINT4 & 10.32 & 64.25 & 74.61 & 41.20 & 67.17 & 42.92 & 58.03 \\
\hline
RTN & 11.03 & 63.46 & 74.10 & 41.00 & 68.48 & 41.81 & 57.77 \\
GPTQ & 10.20 & 64.01 & 75.14 & 39.20 & 70.37 & 42.41 & 58.23 \\
QuaRot & 11.32 & 59.43 & 62.67 & 39.60 & 62.67 & 37.97 & 52.47 \\
QuaRot$^+$ & 9.65 & 63.54 & 75.46 & 38.40 & 70.24 & 43.69 & 58.27 \\
SpinQuant & 9.58 & 63.77 & 75.14 & 40.00 & 72.43 & 44.11 & 59.09 \\
BRQ & 9.48 & 63.38 & 75.52 & 41.80 & 71.97 & 45.73 & 59.68 \\
\hline
\hline
\end{tabular}
\label{tab:qwen3b_quant}
\end{table}

\begin{table}[H]
\centering
\caption{Evaluation of different methods on Qwen2.5 7B across multiple benchmarks.}
\begin{tabular}{cccccccc}
\hline
\hline
\textbf{Method} & \textbf{WiKi} & \textbf{WG} & \textbf{PIQA} & \textbf{OBQA} & \textbf{ARC-E} & \textbf{ARC-C} & \textbf{Avg} \\
\hline
Baseline & 7.81 & 71.35 & 79.43 & 45.40 & 75.72 & 49.74 & 64.33 \\
BINT4 & 9.07 & 67.01 & 77.15 & 43.80 & 73.06 & 46.50 & 61.50 \\
\hline
RTN & 10.00 & 67.80 & 76.01 & 43.60 & 74.79 & 46.93 & 61.83 \\
GPTQ & 9.10 & 65.98 & 77.86 & 44.00 & 75.38 & 47.70 & 62.18 \\
QuaRot & 9.57 & 64.17 & 77.86 & 42.00 & 74.66 & 48.55 & 61.45 \\
QuaRot$^+$ & 8.45 & 65.98 & 76.71 & 42.60 & 74.37 & 49.15 & 61.76 \\
SpinQuant & 8.41 & 66.69 & 78.13 & 43.60 & 76.14 & 49.74 & 62.86 \\
BRQ & 8.34 & 67.96 & 78.40 & 44.80 & 77.02 & 48.89 & 63.41 \\
\hline
\hline
\end{tabular}
\label{tab:qwen7b_quant}
\end{table}

\end{document}